\documentclass[sn-basic]{sn-jnl}

\usepackage{graphicx}%
\usepackage{multirow}%
\usepackage{indentfirst}%
\usepackage{amsmath,amssymb,amsfonts}%
\usepackage{amsthm}%
\usepackage{tabularx}%
\usepackage{mathrsfs}%
\usepackage[title]{appendix}%
\usepackage{xcolor}%
\usepackage{caption}%
\usepackage{subcaption}%
\usepackage{pdflscape}%
\usepackage{textcomp}%
\usepackage{manyfoot}%
\usepackage{booktabs}%
\usepackage{algorithm}%
\usepackage{algorithmicx}%
\usepackage{algpseudocode}%
\usepackage{listings}%

\theoremstyle{thmstyleone}%
\theoremstyle{thmstyletwo}%

\theoremstyle{thmstylethree}%

\begin{document}

\title[Article Title]{Skill-based Explanations for Serendipitous Course Recommendation}


\author*[1]{\fnm{Hung} \sur{Chau}}\email{hkc@pitt.edu}

\author[2]{\fnm{Run} \sur{Yu}}\email{yurun1208@gmail.com}

\author[2]{\fnm{Zachary} \sur{Pardos}}\email{pardos@berkeley.edu}

\author[1]{\fnm{Peter} \sur{Brusilovsky}}\email{peterb@pitt.edu}


\affil*[1]{\orgdiv{Department of Informatics and Networked Systems}, \orgname{University of Pittsburgh}, \orgaddress{\city{Pittsburgh}, \state{PA}, \country{United States of America}}}

\affil[2]{\orgdiv{School of Education}, \orgname{University of California at Berkeley}, \orgaddress{ \city{Berkeley}, \state{CA}, \country{United States of America}}}



\abstract{
Academic choice is crucial in U.S. undergraduate education, allowing students significant freedom in course selection. However, navigating the complex academic environment is challenging due to limited information, guidance, and an overwhelming number of choices, compounded by time restrictions and the high demand for popular courses. Although career counselors exist, their numbers are insufficient, and course recommendation systems, though personalized, often lack insight into student perceptions and explanations to assess course relevance. In this paper, a deep learning-based concept extraction model is developed to efficiently extract relevant concepts from course descriptions to improve the recommendation process. Using this model, the study examines the effects of skill-based explanations within a serendipitous recommendation framework, tested through the AskOski system at the University of California, Berkeley. The findings indicate that these explanations not only increase user interest, particularly in courses with high unexpectedness, but also bolster decision-making confidence. This underscores the importance of integrating skill-related data and explanations into educational recommendation systems.}




\maketitle

\section{Introduction}
\label{chap-intro}

Academic exploration is central to U.S. undergraduate education, yet the elective model can leave students navigating complex choices with limited guidance \cite{Scott-Clayton2015,Goldrick-Rab2006,Four-year-myth}. Existing tools—e.g., Stanford’s CARTA \cite{Chaturapruek2018,Chaturapruek2021} and Pitt’s Degree Planner\footnote{https://www.registrar.pitt.edu/degree-planner}—aggregate useful information (requirements, sequences, outcomes) but do not offer personalized guidance for exploring courses aligned with students’ interests and goals.

Course recommendation has emerged as a remedy, fueled by rich educational data \cite{Ognjanovic2016UsingID,Jiang2019,Fischer2020}. Early systems used social signals or constraints, such as CourseAgent \cite{Farzan2011} and CourseRank \cite{Parameswaran2011}. Recent work leverages deep learning and audit integrations, e.g., AskOski at UC Berkeley \cite{pardos2019connectionist}, and RNN-based preparation recommenders \cite{Jiang2019}. However, human factors remain underexplored: students must weigh opportunity costs, and opaque suggestions can deter them from unfamiliar but valuable areas \cite{Chambliss2014}. Explanations are especially important for serendipity-focused recommenders \cite{pardos2020designing}, where the goal is to surface unexpected yet relevant options that users might otherwise ignore. Explanations can also improve perceived transparency and trust \cite{Bilgic2005,knijnenburg2012inspectability}.

Skills provide a natural explanatory lens. Many content-based methods rely on bag-of-words, which limits both matching quality and interpretability \cite{pardos2020designing,Esteban2018,Jiang2020}. Keyphrases (concepts) communicate semantics more effectively and have been shown to aid understanding and explanation \cite{Liu2007,Tagsplanations2009,Zhang2014,Waern2004,Ahn2007}.

This paper makes two contributions. First, we develop a concept (skill) extraction model for course descriptions, framing extraction as sequence labeling and combining BERT with a BI-LSTM-CRF in a stacking ensemble. Trained on public datasets, the model yields high-quality concepts per expert evaluation, improving over unigram-based representations used previously \cite{Run@2021}. Second, we integrate these skill-based explanations into a serendipitous course recommender built on AskOski\footnote{https://askoski.berkeley.edu/}, powered by PLAN-BERT (an adaptation of BERT4Rec) \cite{Sun2019BERT4Rec}. In a user study, explanations did not change overall ratings of PLAN-BERT’s recommendations, but they significantly increased interest in highly \emph{unexpected} courses under our diversification strategy. Participants reported that explanations were useful, boosted confidence, and reduced neutral responses; effects were pronounced among undeclared students, where the absence of explanations increased neutrality to a statistically significant degree.

Together, these results suggest that skill-based explanations can make serendipitous recommendations more approachable and actionable, particularly for students still exploring their academic paths.

\section{Related Work}
\label{chap-related-work}
\subsection{Automatic Concept Extraction}

Automatic keyphrase extraction has been widely studied across domains like scientific, educational, and biomedical texts, with approaches spanning rule-based, supervised, unsupervised, and deep learning methods. Most systems involve (1) extracting candidate keyphrases via lexical patterns or heuristics, and (2) selecting true keyphrases through ranking or classification. Some recent work frames keyphrase extraction as a sequence-labeling task to better capture semantic dependencies.

\textbf{Feature-based extraction:} Candidates are often selected using POS patterns \cite{Mihalcea2004TextRank,Bougouin2013topic,LiuNAACL09,wan08}, n-gram filtering \cite{Witten1999,Medelyan2009}, Wikipedia matching \cite{Wang2015,Grineva2009}, or noun phrase patterns \cite{FlorescuACL17,Le2016}. Scoring uses features such as term statistics, position, linguistic cues, or external knowledge. Supervised models employ classifiers like logistic regression or SVMs, trained on various feature sets \cite{Hammouda2005,Rose2010rake,Hulth2003,Yih2006,nguyen2017}.

\textbf{Unsupervised approaches:} Graph-based models (e.g., TextRank) score candidates based on graph centrality \cite{Mihalcea2004TextRank,Bougouin2013topic}, while topic clustering groups semantically similar phrases for selection \cite{LiuNAACL09,Grineva2009}.

\textbf{Sequence tagging-based extraction:} Recent work applies sequence labeling (e.g., CRF, Bi-LSTM, BERT) to keyphrase extraction \cite{Bhaskar2012,huang2015bidirectional,Alzaidy2019,Sahrawat2020,Park2020scientific}. These methods leverage pre-trained language models and transfer learning, reducing feature engineering and improving domain transfer. Other advances include combining visual features with deep language models \cite{wang-etal-2020-incorporating,xiong-etal-2019-open}, and generative models like CopyRNN \cite{Meng2017CopyRNN}.

\textbf{Educational concept extraction:} Few works address educational domains, especially course syllabi. Existing projects focus on concept hierarchies \cite{Wang2015}, prerequisite structures \cite{Labutov2017}, or basic extraction using Wikipedia and book indices \cite{Wang2015,Isaac2020,Thaker2020}. Ontology construction often relies on manual or heuristic methods \cite{shamsfard2004learning,Wilson@2012,Zouaq2007,Conde@2014,Conde2016}, but automated extraction remains limited in performance.

FACE \cite{Chau2020} proposes a supervised, feature-rich approach for textbook concept extraction, outperforming previous methods. However, manual annotation for supervised models is costly and impractical for large, unstructured corpora like course syllabi \cite{chau2020understanding}. Deep learning and transfer learning approaches, such as Bi-LSTM-CRF \cite{huang2015bidirectional} and BERT \cite{devlin2019bert}, have shown success in NER and offer promise for concept extraction in course descriptions. In this work, we develop models based on these architectures to address the challenges in educational concept extraction.

\subsection{Course Recommendation}

Course recommendation is a key area within personalized education systems, especially for multidisciplinary programs where students face many choices with limited guidance. Surveys indicate that students’ course selection is driven by factors such as interest, grades, learning goals, career plans, social aspects, and popularity \cite{ma2020,Guruge2021}. The goal of course recommender systems is to provide personalized suggestions based on students' backgrounds and preferences, enhancing learning outcomes and job prospects. However, designing such systems is challenging due to the diversity of student needs, changing curricula, prerequisite structures, and institutional complexity.

Content-based recommenders leverage course features and student profiles (e.g., previous subjects, majors) to suggest relevant courses. Techniques include representing courses as word vectors and matching them to student interests or past enrollments \cite{Esteban2018,GULZAR2018518,Morsomme2019ContentbasedCR,Ma2020CourseRF,Neamah2018}. Some systems use ontologies for domain knowledge, or student queries as proxies for interest, while others utilize bag-of-words or TF-IDF models for course similarity \cite{pardos2020designing,Jiang2020}.

Collaborative filtering approaches predict student preferences by analyzing patterns among similar users. Early works used past performance data and collaborative techniques for grade prediction and course ranking \cite{Sanjog2011,Backenkhler2018DataDrivenAT,Houbraken2017,elbadrawy2016domain}. These models incorporate enrollment history, grade data, and even social or temporal factors, such as graduate attribute ratings or Markov models for sequential recommendations \cite{Bakhshinategh2017ACR,Polyzou2019ScholarsWA}.

Knowledge-based systems combine explicit domain knowledge, ontologies, and user preferences to generate explainable recommendations. Hybrid models integrate content-based, collaborative, and knowledge-based signals to enhance accuracy and personalization \cite{Ibrahim2018,GULZAR2018518,Bydžovská2016,ESTEBAN2020105385,Morsy2019WillTC,pardos2019articulation}.

Other methods include optimization algorithms for course sequencing \cite{Jie2016}, random-walk approaches for sequential dependencies \cite{Polyzou2019ScholarsWA}, and community-based systems leveraging student ratings and analytics \cite{Farzan2011,Chaturapruek2018,Li2012,Parameswaran2011}.

Deep learning-based recommenders have emerged in recent years, utilizing RNNs and transformers to model enrollment sequences and predict suitable courses \cite{Jiang2019,Chris2018,pardos2019connectionist,Shao_Guo_Pardos_2021}. These models improve handling of context, constraints, and the cold start problem, and can incorporate user and item features for personalization.

Despite advancements, few works address human factors such as explainability, novelty, and serendipity—critical for supporting complex, high-stakes student decisions \cite{pardos2020designing}. Existing research also rarely incorporates job information into recommendations and explanations.

\subsection{Explainable Recommendation}

Explainable recommendation is an active research area in explainable AI, aiming to provide not only recommendations but also clear rationales for them. As recommender systems often function as “black boxes,” especially with the adoption of complex models like deep learning, interpretability and transparency are increasingly important for user trust and satisfaction~\cite{Harman2014trust, pardos2019connectionist, PARRA201543}. Early work emphasized transparency and the user’s conceptual understanding of recommendations, laying the groundwork for user-centered explanation models~\cite{Schafer1999, Herlocker2000, Sinha2002}.

Subsequent research in the late 2000s focused on improving the perceived transparency, trust, and acceptance of recommender systems by providing explanations~\cite{wang2007recommendation, tintarev2015explaining, Nunes2017taxonomy, Zhang2020}. Explanations have since been categorized by factors such as reasoning model, display style, and explanatory goals—including transparency, scrutability, trust, persuasiveness, and satisfaction~\cite{Tintarev2007, Zhang2020, gedikli2014should, Sato2018}.

Explanation approaches can be divided into model-intrinsic (interpretable models) and model-agnostic (post-hoc or justification) methods. While early work focused on building inherently interpretable models~\cite{Zhang2014, Herlocker2000}, recent studies have developed post-hoc explanation frameworks for black-box models~\cite{ni-etal-2019-justifying, Musto2020GeneratingPH, Mauro2022}, showing that natural language justifications and simple explanations can improve user trust and engagement.

Most explainable recommendation research has centered on e-commerce, with fewer studies in education. Some educational works have explored pairing reinforcement learning with explanations to improve student engagement~\cite{Zhou2020}, providing explanation generators in quiz recommenders~\cite{Takami2022}, and assessing the impact of explanations on student behavior~\cite{Barria-Pineda2021}. In course recommendation, studies have examined factors influencing student choice~\cite{ma2020} and explored explanation designs, noting that keyphrase-based explanations can improve comprehension over simple unigrams~\cite{Run@2021, Liu2007, Tagsplanations2009, Zhang2014, Waern2004}.

\section{Automatic Concept Extraction for Course Descriptions}\label{sec3}
\label{chap:concept-extraction}

Transfer learning addresses the challenge of adapting models pre-trained on large unannotated datasets to downstream tasks. Recent advances in self-supervised learning with deep neural networks have produced general-purpose models that capture rich structural patterns in data. These models, widely successful in NLP \cite{conneau2017,McCann2017} and computer vision \cite{Yosinski2014}, can be fine-tuned for tasks with limited or weak labels \cite{Shen2022}.

In NLP, CNN-, RNN-, and Transformer-based sequence tagging models leverage pre-trained embeddings and language models to infer semantic and syntactic features, eliminating the need for manual feature engineering. Such models have achieved strong results in Named Entity Recognition (NER) and have been applied to keyphrase extraction \cite{huang2015bidirectional,Med-BERT}.

Concept extraction from course descriptions can be framed as a sequence labeling task akin to NER: given an input sequence $d = {x_1, ..., x_n}$, predict labels $Y = {k_B, k_I, k_O}$ indicating the start, continuation, or absence of a keyphrase.

Domain adaptation seeks to transfer knowledge from a source domain to a related target domain \cite{Wang2018-domain-adaptation,Wilson2020}. In deep learning, this often involves fine-tuning later network layers to capture domain-specific patterns while preserving transferable lower-layer features.

Here, we train concept extraction models on Wikipedia articles, scientific abstracts, and textbook sections, and directly apply them to course descriptions—assuming the source and target domains are sufficiently similar to avoid additional adaptation.

\subsection{Deep Neural Architectures for Concept Extraction}
\subsubsection{Bi-LSTM-CRF}
Bidirectional LSTM-CRF (Bi-LSTM-CRF) models, one of the latest techniques for sequence tagging, are first presented in \cite{huang2015bidirectional}. Different modified versions of the models have been proposed in recent years, models with static word embeddings \cite{lample2016}, models with contextual embeddings \cite{peters2018}, models with additional character embeddings \cite{peters2018,ma2016,Liu2017EmpowerSL}, models adding language model (LM) embeddings \cite{peters2017}, and models with task-aware neural language model \cite{Liu2017EmpowerSL}. Some of these models have been adapted to keyphrase extraction mainly for scientific articles; \cite{Alzaidy2019} uses the Bi-LSTM-CRF model with fixed word embeddings and character embeddings, and \cite{Sahrawat2020} uses the Bi-LSTM-CRF model with contextual embeddings. These studies show the benefits and improvement of keyphrase extraction compared to state-of-the-art unsupervised and supervised models.

The standard Bi-LSTM-CRF model for keyphrase extraction consists of three main components \cite{Liu2017EmpowerSL}: 1. Embedding Layer, 2. Bi-LSTM Layer, and 3. CRF Networks. During inference, CRFs use the Viterbi algorithm to efficiently find the optimal sequence of labels. 

\subsubsection{BERT}
Another widely used deep learning model for keyphrase extraction is Bidirectional Encoder Representations from Transformers (BERT). BERT \cite{devlin2019bert} is a pre-trained deep neural network that employs multi-layer bidirectional Transformers to learn contextualized language representations from large volumes of unlabeled text. By fine-tuning these pre-trained representations, BERT has achieved state-of-the-art results across various natural language processing (NLP) tasks, including language understanding, text classification, question answering, and named entity recognition (NER). Its ability to model the contextual meaning of words and phrases makes it particularly effective for keyphrase extraction \cite{Med-BERT,Wang2020}.

Fine-tuning BERT for a downstream task leverages the Transformer’s self-attention mechanism, which enables the model to capture dependencies across tokens regardless of their distance in the text. This flexibility allows BERT to handle a variety of task formats—ranging from single-sentence classification to sentence-pair modeling—by appropriately configuring the input and output layers. For token-level tasks such as sequence tagging, the token embeddings produced by BERT’s final layer are passed to an additional classification layer.

In this work, we adopt the BERT architecture for concept extraction from course descriptions. Specifically, the token representations from BERT’s output layer serve as input to a token classification module that assigns each token to one of three classes: \textit{O}, \textit{B-CON}, or \textit{I-CON}.

\subsection{Experiments and Results}
\subsubsection{Training Datasets}
To overcome the problem of manually labeling data for course descriptions, we apply the notion of domain adaption to train the concept extraction models on source datasets close to the target course description, with expert-annotated labels or weak labels. Specifically, several existing labeled datasets in the academic and educational domains will be used to train models to extract concepts for course descriptions.

\textbf{Introduction to Information Retrieval (IIR) dataset}\footnote{https://github.com/PAWSLabUniversityOfPittsburgh/Concept-Extraction}: contains a section-level concept index for the first 16 chapters of the book ``Introduction to Information Retrieval'' (IIR) \cite{Wang2021}.

\textbf{KP20K}\footnote{https://github.com/memray/seq2seq-keyphrase}: one of the largest datasets in scientific keyphrase studies developed by Rui at el. \cite{Meng2017CopyRNN}. It contains the titles, abstracts, and keyphrases of 554,133 scientific articles in the Computer Science domain.

\textbf{Wikipedia}: the largest dataset in this study. It contains Wikipedia articles including page topics which can be used to filter data to train domain-specific concept extractors. We process 636,917 articles to train models to extract concepts in Computer and Information-related course descriptions.

\subsubsection{Implementation Details}

Various models are trained with Bi-LSTM-CRF and BERT architectures using three different training datasets separately and two training settings (i.e., \textit{uncased} texts and \textit{cased} texts). In total, there are six main models trained for each architecture, plus a combined model that combines the outputs of the six models together:

\begin{itemize}
    \item \textit{uncased-iir}: model trained with \textit{IIR} dataset in the \textit{uncased} setting
    \item \textit{cased-iir}: model trained with \textit{IIR} dataset in the \textit{cased} setting
    \item \textit{uncased-kp20k}: model trained with \textit{KP20K} dataset in the \textit{uncased} setting
    \item \textit{cased-kp20k}: model trained with \textit{KP20K} dataset in the \textit{cased} setting
    \item \textit{uncased-wiki}: model trained with \textit{Wikipedia} dataset in the \textit{uncased} setting
    \item \textit{cased-wiki}: model trained with \textit{Wikipedia} dataset in the \textit{cased} setting
    \item \textit{combined-all}: model combines the output of each of the above models together
\end{itemize}

All the trained models are available on this Github repository.\footnote{https://github.com/HungChau/course-concept-extraction} The implementation of Bi-LSTM-CRF models is based on the version\footnote{https://github.com/LiyuanLucasLiu/LM-LSTM-CRF} presented in \citep{Alzaidy2019} at the sentence level. The character embeddings of 30 dimensions are obtained by training additional Bi-LSTM networks along with the main model. For the word embeddings, the Glove pre-trained word embeddings of 100 dimensions\footnote {https://nlp.stanford.edu/projects/glove/} are used.  A 300-dimensional hidden layer of LSTM units is used for both the character-level embedding model and the main model. The models are trained using mini-batch stochastic gradient descent with momentum. The batch size, learning rate and decay ratio are set to 10, 0.015 and 0.05, respectively. The dropout strategy is also applied to avoid over-fitting and gradient clipping of 5.0 to increase the model's stability. For BERT models, a distilled version (DistilBERT) is used. It is smaller, faster, cheaper and lighter, yet achieves competitive performances compared to the original architecture \cite{distilbert}.

\subsection{Model Performances}
To evaluate and compare the performance among the models, we created a test set including 50 randomly selected course descriptions. we manually annotate concept labels for each of the course descriptions. The trained models are evaluated using standard keyphrase extraction metrics on this set of 50 course descriptions.

Table \ref{tab:concept-extraction-perf} shows the performance of two different models, BERT and BI-LSTM-CRF, for concept extraction. The performance is measured in terms of precision, recall, and F1 score for three different training datasets (i.e., IIR, KP20K and Wikipedia) and two training settings (i.e., \textit{case} and \textit{uncased}). The \textit{combined-all} row represents the performance of the stacking ensemble models which combines the outputs of the six based models for each of the deep architectures separately. The last row of the table shows the performance of all BERT and BI-LSTM-CRF models combined on a test set. The best performing model for each dataset and metric (or overall) is highlighted in bold.

As can be seen from Table \ref{tab:concept-extraction-perf}, in terms of F1 score,  BERT demonstrates superior performance on IIR and KP20K datasets, while BI-LSTM-CRF performs better on the Wikipedia dataset. The stacking ensemble model exhibits the best performance in terms of F1 score, regardless of the architecture employed, with the BERT ensemble model surpassing the BI-LSTM-CRF ensemble model. The best performance across all three metrics, namely precision (0.758), recall (0.625), and F1 score (0.685), is achieved by combining the BERT and BI-LSTM-CRF models.

\begin{table}[]
\small
\centering
\caption{Model performance summary of BERT and BI-LSTM-CRF on a task of concept extraction for course descriptions.}
\begin{tabular}{|l|llllll|}
\hline
                       & \multicolumn{3}{c|}{\textbf{BERT}}                                                                              & \multicolumn{3}{c|}{\textbf{BI-LSTM-CRF}}                                                  \\ \hline
                       & \multicolumn{1}{l|}{precision}      & \multicolumn{1}{l|}{recall}         & \multicolumn{1}{l|}{f1}             & \multicolumn{1}{l|}{precision}      & \multicolumn{1}{l|}{reall}          & f1             \\ \hline
\textit{uncased-iir}   & \multicolumn{1}{l|}{\textbf{0.738}} & \multicolumn{1}{l|}{0.269}          & \multicolumn{1}{l|}{0.394}          & \multicolumn{1}{l|}{0.773}          & \multicolumn{1}{l|}{0.256}          & 0.385          \\ \hline
\textit{cased-iir}     & \multicolumn{1}{l|}{0.726}          & \multicolumn{1}{l|}{0.281}          & \multicolumn{1}{l|}{0.405}          & \multicolumn{1}{l|}{\textbf{0.844}} & \multicolumn{1}{l|}{0.208}          & 0.334          \\ \hline
\textit{uncased-kp20k} & \multicolumn{1}{l|}{0.515}          & \multicolumn{1}{l|}{0.195}          & \multicolumn{1}{l|}{0.283}          & \multicolumn{1}{l|}{0.714}          & \multicolumn{1}{l|}{0.111}          & 0.193          \\ \hline
\textit{cased-kp20k}   & \multicolumn{1}{l|}{0.537}          & \multicolumn{1}{l|}{0.217}          & \multicolumn{1}{l|}{0.309}          & \multicolumn{1}{l|}{0.570}          & \multicolumn{1}{l|}{0.148}          & 0.236          \\ \hline
\textit{uncased-wiki}  & \multicolumn{1}{l|}{0.629}          & \multicolumn{1}{l|}{0.259}          & \multicolumn{1}{l|}{0.367}          & \multicolumn{1}{l|}{0.797}          & \multicolumn{1}{l|}{0.285}          & 0.420          \\ \hline
\textit{cased-wiki}    & \multicolumn{1}{l|}{0.608}          & \multicolumn{1}{l|}{0.248}          & \multicolumn{1}{l|}{0.352}          & \multicolumn{1}{l|}{0.809}          & \multicolumn{1}{l|}{0.313}          & 0.452          \\ \hline
\textit{combined-all}  & \multicolumn{1}{l|}{0.733}          & \multicolumn{1}{l|}{\textbf{0.556}} & \multicolumn{1}{l|}{\textbf{0.633}} & \multicolumn{1}{l|}{0.799}          & \multicolumn{1}{l|}{\textbf{0.503}} & \textbf{0.617} \\ \hline
                       & \multicolumn{6}{c|}{\textbf{BERT + BI-LSTM-CRF}}                                                                                                                                                             \\ \hline
                       & \multicolumn{2}{l|}{precision}                                            & \multicolumn{2}{l|}{recall}                                               & \multicolumn{2}{l|}{f1}                              \\ \hline
\textit{deep-concept-extractor}  & \multicolumn{2}{l|}{\textbf{0.758}}                                       & \multicolumn{2}{l|}{\textbf{0.625}}                                       & \multicolumn{2}{l|}{\textbf{0.685}}                  \\ \hline
\end{tabular}
\label{tab:concept-extraction-perf}
\end{table}

Here is an example of concepts extracted from an actual Algorithm course at Pitt. The yellow-highlighted phrases represent the concepts extracted by the \textit{combined} model. 

``\textit{This course emphasizes the study of the basic \colorbox{yellow}{data structures} of \colorbox{yellow}{computer science} (\colorbox{yellow}{stacks}, \colorbox{yellow}{queues}, \colorbox{yellow}{trees}, \colorbox{yellow}{lists}) and their implementations using the \colorbox{yellow}{java language} included in this study are \colorbox{yellow}{programming techniques} which use \colorbox{yellow}{recursion}, \colorbox{yellow}{reference variables}, and \colorbox{yellow}{dynamic memory allocation}.  Students in this course are also introduced to various \colorbox{yellow}{searching} and \colorbox{yellow}{sorting methods} and also expected to develop an intuitive understanding of the \colorbox{yellow}{complexity} of these \colorbox{yellow}{algorithms}...}''

\subsection{Expert Evaluation}
In addition to the offline evaluation on a test set, we also conduct an expert evaluation to ensure the quality of extracted concepts for course recommendation applications. we hire two PhD students who specialize in Computing and Information Science. To begin the evaluation, we randomly sample 50 course descriptions in SCI. For each description, we apply the trained model to extract concepts in the text. The two experts are then provided with the course descriptions and a list of extracted concepts for each course. The experts are asked to rate each extracted concept as either good or not good for the corresponding course. Table \ref{tab:testset-stat} summarizes the statistics of the evaluation dataset.

\begin{table}[]
\small
\centering
\caption{Expert evaluation dataset statistics}
\begin{tabular}{|l|l|}
\hline
Number of course descriptions                        & 50    \\ \hline
Average number of words per description              & 72.0  \\ \hline
Average number of extracted concepts per description & 12.64 \\ \hline
Number of extracted concepts                         & 632   \\ \hline
Number of unique extracted concepts                  & 519   \\ \hline
\end{tabular}
\label{tab:testset-stat}
\end{table}

Table \ref{tab:expert-eval} presents the results of the expert evaluation conducted to measure the performance of the \textit{deep-concept-extractor} model. The table reports four metrics: macro accuracy, micro accuracy, proportional agreement, and Kappa agreement. Macro accuracy measures the accuracy of the experts' evaluations at a high level for all the course descriptions together, while micro accuracy measures the average accuracy at an individual course level. The results show that both experts have high levels of agreement with the concepts extracted by the model, with macro accuracy ranging from 90.51\% to 90.98\%, and micro accuracy ranging from 89.54\% to 90.43\%. Furthermore, the proportional agreement between the two experts was 92.88\%,  indicating that they agreed on their assessments most of the time. Finally, the table reports the Kappa agreement between the two experts, which measures the level of agreement between them beyond what would be expected by chance. The Kappa agreement between both experts was 0.57, indicating a good level of agreement. Overall, the results of the expert evaluation illustrate that the two experts frequently agree with the outputs of the concept extraction model. Consequently, the combined BERT and BI-LSTM-CRF models will be employed to extract concepts from descriptions for explainable course recommendation systems in the next sections.

\begin{table}[]
\small
\centering
\caption{The result of the expert evaluation.}
\begin{tabular}{|l|c|c|c|}
\hline
\textbf{Metric}        & \textbf{Expert 1} & \textbf{Expert 2} & \textbf{Both Experts} \\ \hline
Macro accuracy         & 90.98\%           & 90.51\%           & 87.18\%               \\ \hline
Micro accuracy         & 90.43\%           & 89.54\%           & 86.09\%               \\ \hline
Proportional agreement & --                & --                & 92.88\%               \\ \hline
Kappa agreement        & --                & --                & 0.57                  \\ \hline
\end{tabular}
\label{tab:expert-eval}
\end{table}

\section{Skill-based Explanations for Serendipitous Course Recommendation}
\label{chap-serenditious-course-rec}

The serendipitous course recommendation system aims to recommend courses that are unexpected or novel yet still relevant. The underlying hypothesis is that students are more likely to accept such recommendations. Yet, this task presents significant challenges, particularly within a university setting. In this context, relevant but unexpected courses may belong to departments outside of a student's primary field of study, and their course descriptions may employ unfamiliar terminology, potentially making them less likely to be adopted by students. Previous research, as demonstrated in our study \cite{Run@2021}, underscores the effectiveness of catering to students' prior knowledge by providing personalized explanations. However, it is posited that the efficacy of these explanations can be further enhanced by using keyphrases instead of unigrams. Unigrams, single words devoid of context, may struggle to convey the full meaning encapsulated in course descriptions, particularly in cases where technical terminology is utilized. Keyphrases, on the other hand, have shown promise in improving user comprehension over unigrams, as they are better equipped to communicate the underlying semantics \cite{Waern2004,Ahn2007,Bakalov2013,Glowacka2013}.

In this context, we hypothesize that augmenting course recommendations with skill-based explanations could substantially enhance measures within higher education course recommender systems. Specifically, we propose that by furnishing students with comprehensive information about a course, including how it aligns with their prior knowledge and the novel knowledge it offers, students will be better equipped to evaluate its relevance, be more confident in making decisions and be less likely to dismiss it based on unfamiliarity. We conduct an online user study at the University of California, Berkeley, leveraging the AskOski system powered by PLAN-BERT, an adaptation of BERT4Rec, which is a state-of-the-art deep neural network model for top-N recommendation \cite{Sun2019BERT4Rec}.

\subsection{Recommendation Method}

\begin{figure*}[hbt!]
\centering
 \includegraphics[width=1.0\textwidth]{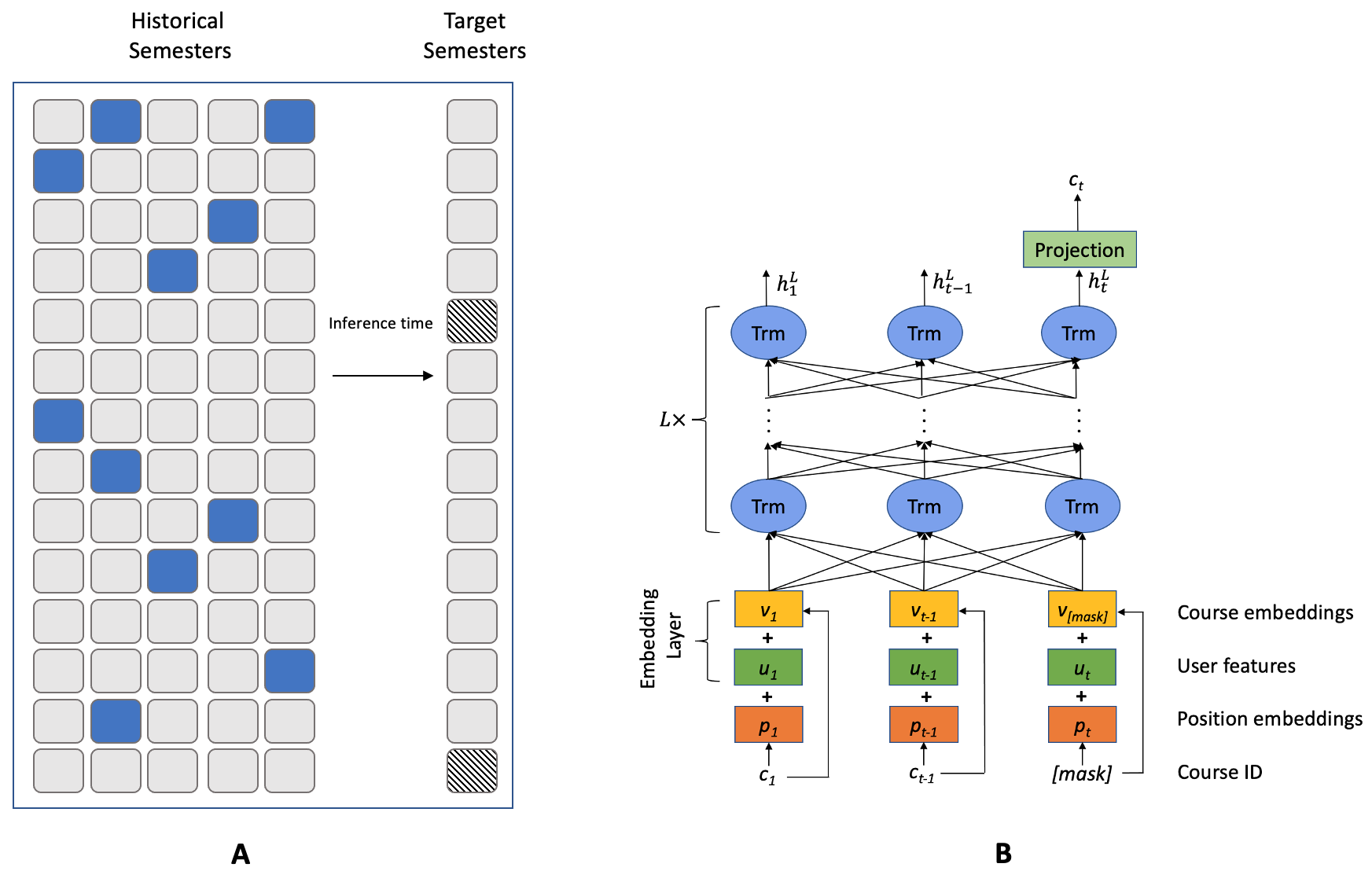}
 \caption{A) Student course enrollment history for training PLAN-BERT: before the inference time is the input and courses after the inference time are masked as the prediction targets; each column represents a semester in the student enrollment history; blue cells represents enrolled courses; and striped cells are enrolled courses in the latest historical semester and masked for prediction. B) BERT architecture for next course prediction task using student course enrollment histories and major information. The position embeddings can be encoded as relative semesters elapsed since the student began.}
 \label{fig:bert2rec}
\end{figure*}

In the realm of recommendations, course recommendation for exploration could be considered as the top \textit{N} item recommendation problem. Early works on this type of recommender system predominantly utilized content-based or collaborative-based approaches. Some of these models leverage course representation models (e.g., \textit{skill}-based or course2vec), as discussed in previous sections, to generate recommendations. To consider time or order information, state-of-the-art technologies employ deep sequential models such as Recurrent Neural Networks or transformer-based models like BERT. Deep learning has gained a considerable amount of interest in numerous research areas such as natural language processing (NLP), speech recognition and computer vision. This surge in interest is not only due to their remarkable performance but also due to their inherent capability of transfer learning and crafting feature representations from scratch. Deep neural networks are also composite in the sense that multiple neural building blocks can be composed into a single big differentiable function and trained in an end-to-end manner. This is the key advantage when coping with content-based recommendations and inevitable when modeling users and items, where multi-modal data is ubiquitous. Deep learning models have successfully been applied to top \textit{N} recommendation problem and shown to be superior over the traditional approaches. Consequently, to curate a list of potential courses to recommend to students based on their course enrollment history, we use BERT4Rec \cite{Sun2019BERT4Rec}, a leading model for top-\textit{N} recommendations. PLAN-BERT  \cite{Shao_Guo_Pardos_2021} (adapted from BERT) has recently been studied and evaluated for multi-semester course recommendation. 
Its bidirectional self-attention design is more effective at utilizing past sequence information compared to both BiLSTM and a UserKNN baseline. Additionally, PLAN-BERT incorporates student characteristics (such as major, division, and department) and course attributes (such as subject and department) to enhance personalization and enhance the quality of recommendations. The online study demonstrated that PLAN-BERT has practical potential to assist students as they navigate the complexities of higher education. Consequently, we use a modified version of PLAN-BERT (refer to Figure \ref{fig:bert2rec}) to generate a list of course suggestions for a specific semester.

\textbf{Training Process:} in order to efficiently train the BERT architecture with sequential data, we will apply the Masked Language Model, a masking technique in NLP. We will employ percentage sampling to pre-train PLAN-BERT to learn the contextual embeddings of courses by predicting the original IDs of the masked courses based only on their left and right contexts. The input format for BERT to learn course embeddings, for example, is:

\begin{center}
{$[c_1, c_2, c_3, c_4, c_5, c_6, c_7] \xrightarrow[mask]{randomly} [c_1, [mask]_1, c_3, c_4, c_5, [mask]_2, c_7]$}
\end{center}

At the main training stage, we will fine-tune PLAN-BERT on the next item prediction task. The courses in the latest historical semester will be masked for prediction (see Figure \ref{fig:bert2rec}.A). In this way, the trained model can learn the relationship between past and future courses.

\textbf{Diversifying Recommendations:} Among the institutional values of a large multidisciplinary university is to expose students to not solely complementary knowledge but also different viewpoints expressed through courses. It helps students expand their learning and collaboration and experience various intellectual schools of thought across the university.  The goal of our course recommendation is to help students improve awareness of course options and explore courses they may find interesting but which have been relatively unexplored by those with similar course selections to them in the past. To achieve this goal and counteract the filter bubble issue of collaborative filtering based recommendation models, we diversify course suggestions by allowing only one result per department \cite{pardos2020designing,Run@2021}, produced by PLAN-BERT.

\subsection{Explanation Method}
One of the primary objectives of this study is to investigate how providing explanations or justifications impacts and enhances user responses to the recommendations generated by the method described in the previous section. Preliminary results from a prior study \cite{Run@2021} indicated that presenting students with prior knowledge (i.e., skills shared between the suggested course and the courses the student has taken) was an effective method for generating personalized explanations that led to increased average ratings across all outcome measures. Furthermore, presenting multi-gram skills instead of unigrams in explanations could potentially further enhance the recommendation outcomes. Unigrams may not have sufficient capacity to convey the nuanced meanings encapsulated in course descriptions. It can be challenging for students to interpret the meaning of individual words, especially technical terms. Therefore, we will utilize multi-gram concepts, extracted by the trained model presented in Section \ref{chap:concept-extraction}, to provide these explanations.

\textbf{The methodology.} Our approach to providing explanations consists of two key aspects: (1) establishing connections between the target course recommendation and skills from courses the student has previously taken, and (2) unveiling novel skills that are taught in the target course. Consequently, the explanation for a recommended course will consist of two separate lists of skills, offering the student both familiar knowledge they have already acquired and new knowledge they have yet to encounter. The two lists of skills are defined as follows:

\begin{equation}
\begin{aligned}
Learned\_Skills = S_t \cap\ (\bigcup_{c\in C}S_c)
\end{aligned}
\end{equation}

\begin{equation}
\begin{aligned}
New\_Skills = S_t - (\bigcup_{c\in C}S_c)
\end{aligned}
\end{equation}

Where $S_t$ is the set of extracted skills for the target recommended course's description, $S_c$ is the set of extracted skills for a taken course's description, and $C$ is the list of courses the student has taken in the past.

\textbf{Skill ranking:} To assess the relevance of an extracted skill to the target course, I compute the relationship, denoted as $r_c(s)$, between the skill and the course. First, each skill and course description are represented as embedding vectors with 768 dimensions, utilizing the \textit{all-mpnet-base-v2} version of SBERT. Subsequently, I calculate the relationship ($0 <= r_c(s) <= 1$) using cosine similarity. This cosine score is employed to rank the list of skills, with the top \textit{N} skills selected to construct the explanation.

\textbf{Skill matching:} two skills could be semantically equivalent but may have different word forms (e.g., \textit{K-Means Clustering} and \textit{K-Means Algorithm}). Exact string matching could lead to missing overlapping skills between the \textit{target} course and a \textit{taken} course. Therefore, I employ a \textit{soft} matching approach, utilizing cosine similarity between two embedding vectors representing the two skills. These embedding vectors are derived from SBERT. Two skills are considered a match if their cosine similarity ($r(s_1, s_2)$) exceeds \textit{0.85}. Note that this heuristic threshold is chosen based on experimental analysis with the SBERT's embeddings. This threshold could be relaxed depending on specific applications or adjusted for different types of embeddings.

\subsection{Study Experiments}

This section outlines the implementation of the proposed skill-based explanation for serendipitous course recommendation at at the University of California, Berkeley. To assess its effectiveness, we conducted a online user study involving undergraduate students, soliciting their insights on multiple dimensions. The study aims to empirically investigate the hypothesis that enhancing course recommendations with explanations can empower students to more effectively assess the relevance of suggested courses. This augmentation is anticipated to reduce the prevalence of neutral opinions and mitigate the likelihood of students disregarding recommendations due to unfamiliarity.

\subsubsection{Implementation Details}

PLAN-BERT underwent retraining using enrollment history data up to Fall 2022. From the initial pool of 20,282 courses, we excluded 553 courses with inadequate descriptions (i.e., fewer than 7 words), resulting in a refined collection of 20,729 courses available for recommendations across all majors. Using student enrollment data and major information as inputs, PLAN-BERT ranks these courses for each student's recommendation. The final recommendations comprise the top 5 courses from 5 distinct departments, ensuring a diverse selection for students. The final five recommended courses are shown to participants in a random order.

To extract the skills encapsulated within these courses, we employed the concept extraction model detailed in Section \ref{chap:concept-extraction}. This process entailed post-processing, including the removal of generic skills such as `homework', `student' and `seminar', skills containing more than 5 words, as well as the consolidation of singular and plural forms of skills. After post-processing, on average, there are 6.5 extracted skill per course.

Both individual skills and course descriptions are transformed into embedding vectors with 768 dimensions, utilizing the \textit{all-mpnet-base-v2} version of SBERT. Subsequently, we calculated the relationships between skills and courses ($0 \leq r_c(s) \leq 1$) via cosine similarity. This cosine score is used to rank the list of skills associated with the course.

To provide insightful explanations for each course recommendation, we employ the skill matching methodology described earlier, comparing the target course with the courses the student has previously taken. This approach identifies the top 7 acquired skills and the top 7 new skills,\footnote{The actual number of skills shown to the subject may be less than 7, depending on how many skills are in the course and how many skills are matched.} which are then presented to the student alongside the course recommendations, offering valuable insights and aiding in informed decision-making.

\subsubsection{User Study}
\begin{figure*}
\centering
 \includegraphics[width=1.0\textwidth]{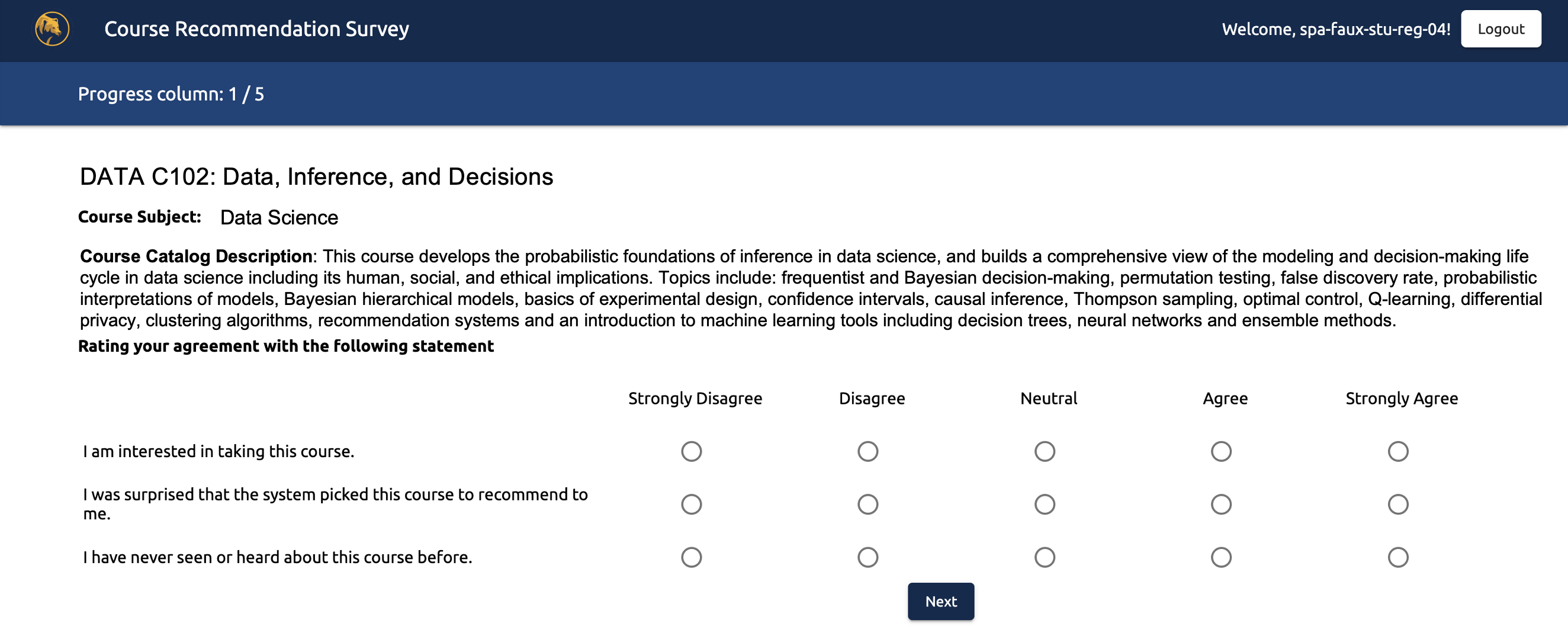}
 \caption{A demonstration of a recommended item with no explanation (group C1) through the AskOski system.}
 \label{fig:noexp_survey_berkeley}
\end{figure*}

\begin{figure*}
\centering
 \includegraphics[width=1.0\textwidth]{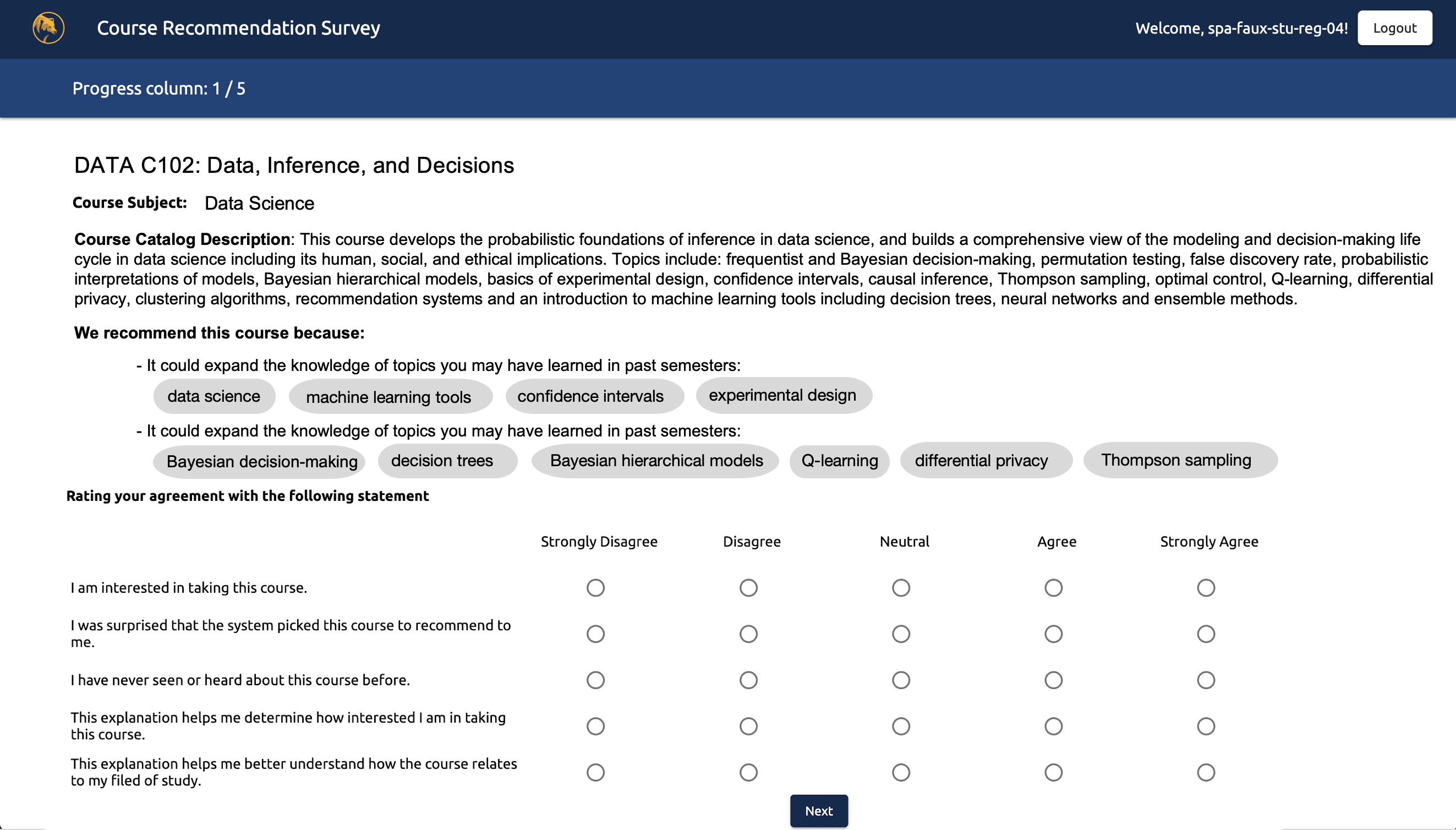}
 \caption{A demonstration of a recommended course for with skill-based explanation (group C2) through the AskOski system. The explanation shows the top 7 learned concepts as well as the top 7 novel concepts offered by the course.}
 \label{fig:exp_survey_berkeley}
\end{figure*}

\textbf{Procedures.} This study is conducted online through the AskOski system at the University of California, Berkeley. Student participants are required to enroll for a minimum of two semesters to take part in the study. They are randomly assigned to one of the two \textit{between}-subject conditions (\textit{Explanation}). The study begins with participants logging into the AskOski system using their Berkeley credentials. Upon accessing the system, participants are presented with the study's introduction. Subsequently, they are presented with a curated list of five course recommendations. These recommendations are tailored based on the participant's past course history, major information, as well as the course history of ``similar'' students. Each course recommendation includes the course ID and title, the description of the course, skill-based explanation (available only for participants in the \textit{Explanation} conditions), and a survey questionnaire consisting of multiple-choice questions (see Fig. \ref{fig:noexp_survey_berkeley} and \ref{fig:exp_survey_berkeley}). Participants are requested to thoroughly review each of the five course recommendations and respond to a series of questions regarding their preferences and feedback for each recommended course. In total, each participant is expected to evaluate and provide feedback on all five recommended courses.

\textbf{Participants.} A total of 53 participants were recruited for this study through a combination of social media and email advertisements. Eligibility for participation was limited to undergraduate students at the University of California, Berkeley, who had completed a minimum of two semesters. Among the recruited participants, 28 were assigned to the explanation condition, and 25 were in the no-explanation one.
These participants come from a diverse array of academic backgrounds, encompassing fields such as Computer Science, Data Science, Economics, Statistics, and Media Studies, among others. The study was designed to be completed within 30 - 45 minutes. Each participant received a \$20 Amazon gift card upon the successful completion of the study.

\textbf{Design and Analysis.} The study is designed as a \textit{between}-subjects study to measure the effect of explanation on the serendipitous course recommendation w.r.t \textit{success}, \textit{unexpectedness} and \textit{novelty}. There are two \textit{between}-subject conditions (\textit{Explanation}): \textit{No-Exp} (C1) vs. \textit{Exp} (C2).

We collected the following measures:
\begin{itemize}[leftmargin=.5in]

    \item Q1. Success (Interest): Participants respond to the statement ``\textit{I am interested in taking this course.}'' \cite{Pu2011,pardos2020designing,Run@2021} on a 5 point Likert scale from 1=Strongly Disagree to 5=Strongly Agree (see Fig. \ref{fig:noexp_survey_berkeley}).
    \item Q2. Unexpectedness: Participants respond to the statement ``\textit{I was surprised that the system picked this course to recommend to me.}'' \cite{Kotkov2018} on a 5 point Likert scale from 1=Strongly Disagree to 5=Strongly Agree (see Fig. \ref{fig:noexp_survey_berkeley}).
    \item Q3. Novelty: Participants respond to the statement ``\textit{I have never seen or heard about this course before.}'' \cite{Pu2011} on a 5-point Likert scale from 1=Strongly Disagree to 5=Strongly Agree (see Fig. \ref{fig:noexp_survey_berkeley}).
    \item Q4. Explanation Effectiveness: Participants respond to the statement ``\textit{This explanation helps me determine how interested I am in taking this course.}'' \cite{vig2009tagsplanations} on a 5 point Likert scale from 1=Strongly Disagree to 5=Strongly Agree (see Fig. \ref{fig:exp_survey_berkeley}).
    \item Q5. Usefulness of Concepts: Participants respond to the statement ``\textit{The explanation helps me better understand how the course relates to my field of study.}'' on a 5-point Likert scale from 1=Strongly Disagree to 5=Strongly Agree (see Fig. \ref{fig:exp_survey_berkeley}).

\end{itemize}

In our study, each participant evaluate several recommended items. Considering repeated measurements made by the same participant as independent could potentially result in a violation of correlated errors \cite{Bart2013,Knijnenburg2015}. To tackle this problem, we employ Generalized Linear Mixed Models for the analyses. These models consider that the ratings are provided by the same users, treating them as random effects, and permitting the estimation of error correlations stemming from the repeated measurements.

\subsubsection{Results}
\begin{figure*}
  \begin{subfigure}{0.5\textwidth} 
    \centering
    \includegraphics[width=\linewidth]{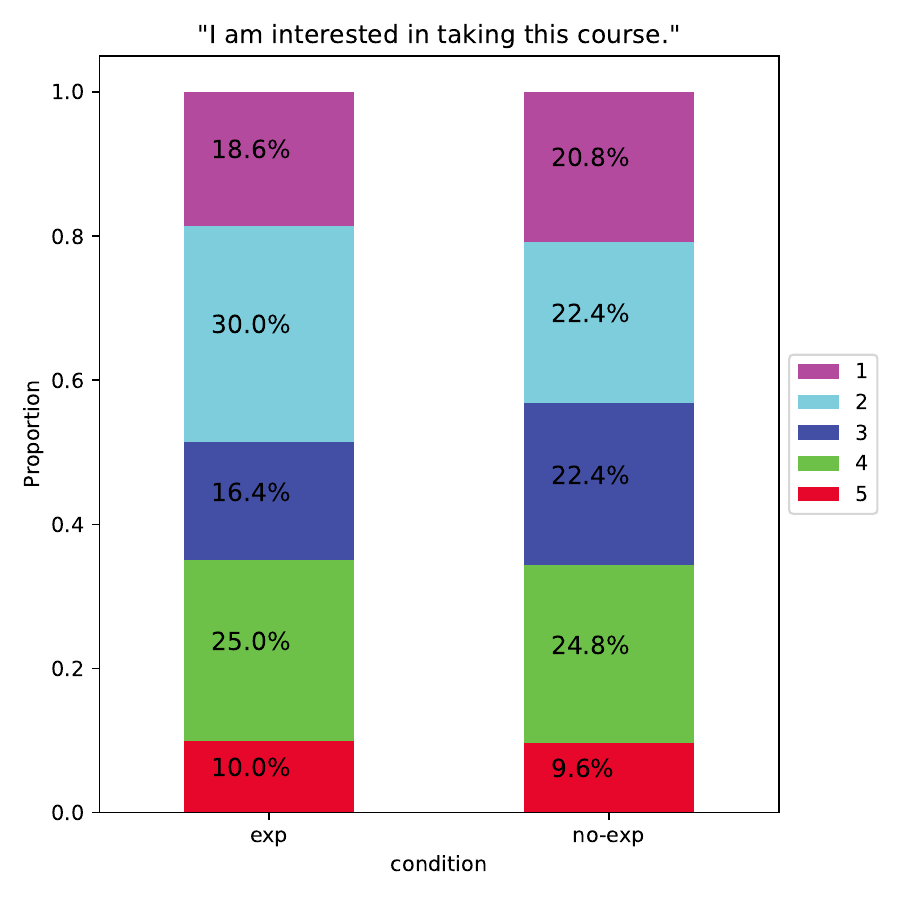}
    \caption{}
    \label{fig:q1-stacked-bar-berkeley}
  \end{subfigure}%
  \begin{subfigure}{0.5\textwidth} 
    \centering
    \includegraphics[width=\linewidth]{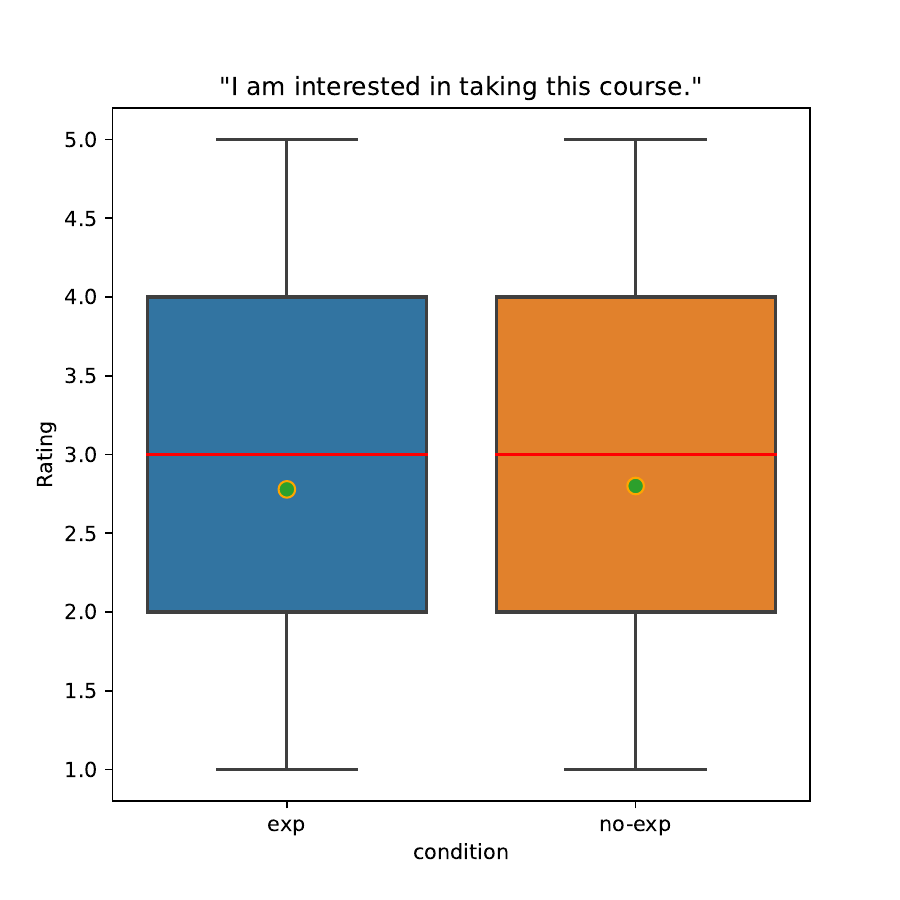}
    \caption{}
    \label{fig:q1-boxplot-berkeley}
  \end{subfigure}
  \caption{(a) Proportional distribution of user responses to the statement `I am interested in taking this course.', comparing those with \textit{Explanation} (exp) and  \textit{Without Explanation} (no-exp). Ratings: 1 - `Strong Disagree', 2 - `Disagree', 3 - `Neutral', 4 - `Agree', 5 - `Strong Agree'. (b) A graph displaying the distribution of ratings in response to research question Q1 for the two conditions, with the median indicated by red lines and the average represented by green circles.}
  \label{fig:q1-berkeley}
\end{figure*}

\textbf{Interestedness.} When comparing the \textit{baseline} model, representing only the intercept, with the \textit{random intercept} model, which accounts for variations among different participants, the findings revealed statistically significant disparities in intercepts across participants. Consequently, we incorporated \textit{participant} random effects into the main analysis. As illustrated in Figure \ref{fig:q1-berkeley}, it is generally observed that participants in the \textit{Exp} conditions (\textit{M} = 2.78, \textit{N} = 140) display slightly less interest in enrolling in the recommended courses compared to those in the \textit{No-Exp} conditions (\textit{M} = 2.8, \textit{N} = 125). Our statistical analysis further indicates that there is a significant variation in intercepts across participants concerning the relationship between the provision of an explanation and participants' interest in taking the course, \textit{SD} = 0.42 (95\% CI: 0.25, 0.71), $=\chi^2(1) = 5.108, p = .02$. However, the explanation itself does not appear to have a significant effect on participants' level of interest in enrolling in the course, \textit{b} = -0.021, \textit{t}(51)=0.11, \textit{p} = 0.91.

\begin{figure*}
  \begin{subfigure}{0.5\textwidth} 
    \centering
    \includegraphics[width=\linewidth]{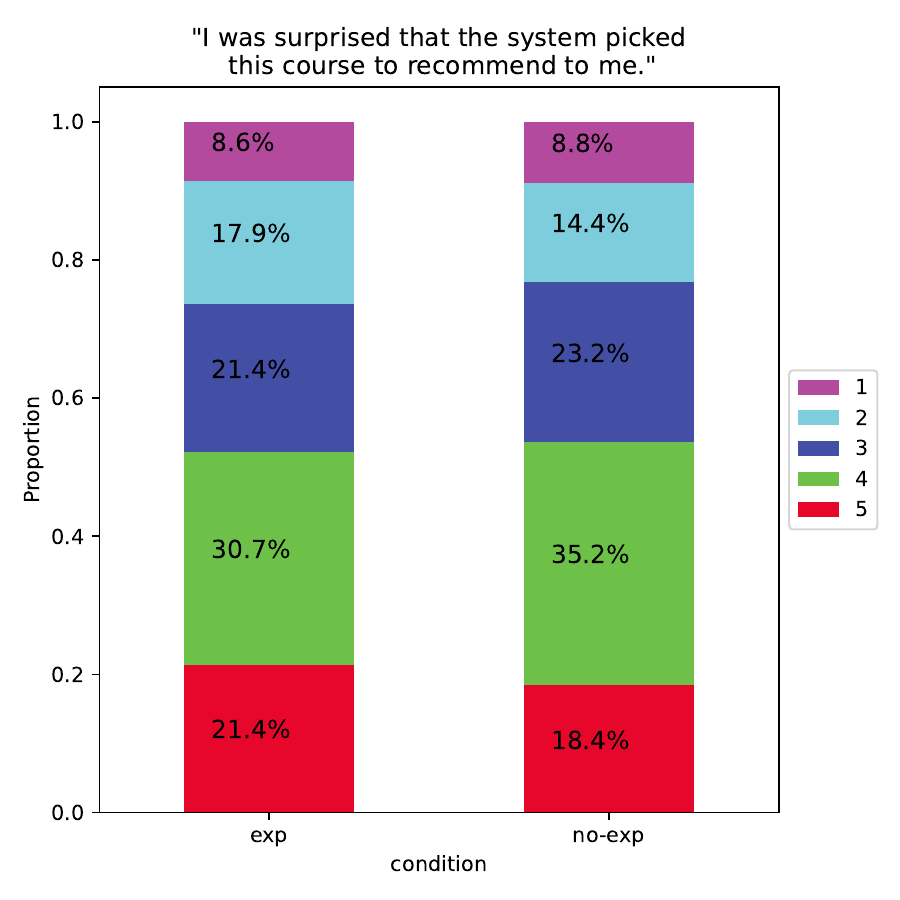}
    \caption{}
    \label{fig:q2-stacked-bar-berkeley}
  \end{subfigure}%
  \begin{subfigure}{0.5\textwidth} 
    \centering
    \includegraphics[width=\linewidth]{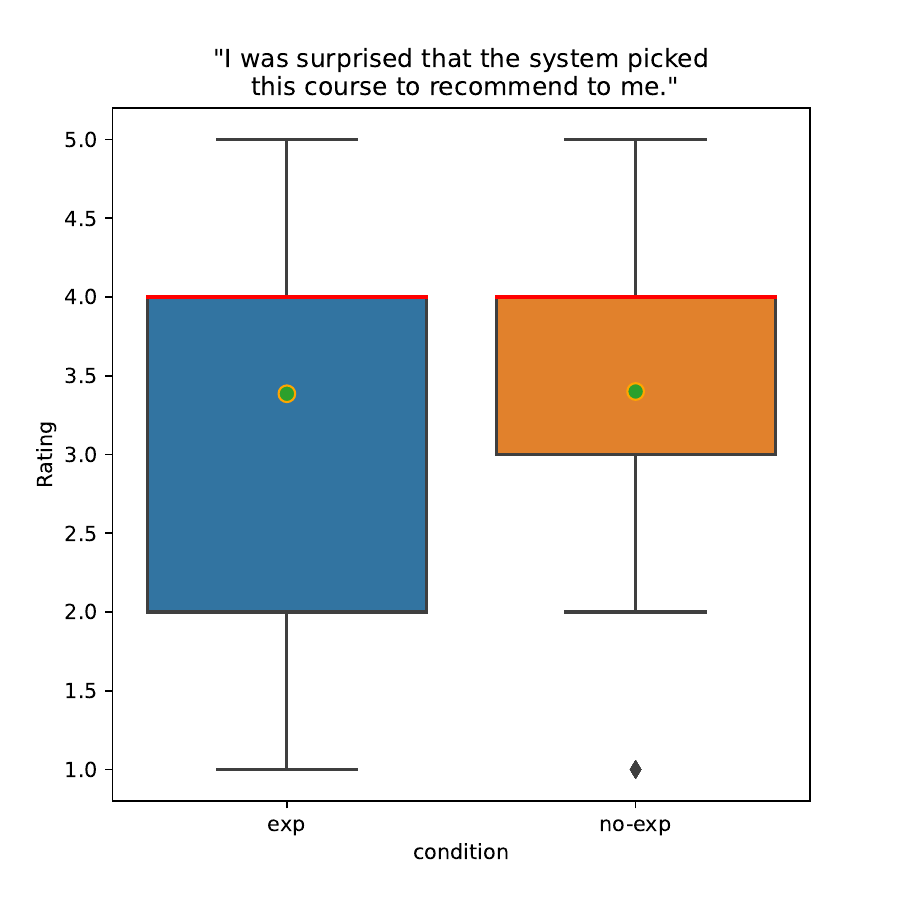}
    \caption{}
    \label{fig:q2-boxplot-berkeley}
  \end{subfigure}
  \caption{(a) Proportional distribution of user responses to the statement `I was surprised that the system picked this course to recommend to me.', comparing those with \textit{Explanation} (exp) and  \textit{Without Explanation} (no-exp). Ratings: 1 - `Strong Disagree', 2 - `Disagree', 3 - `Neutral', 4 - `Agree', 5 - `Strong Agree'. (b) A graph displaying the distribution of ratings in response to question Q2 for the two conditions, with the median indicated by red lines and the average represented by green circles.}
  \label{fig:q2-berkeley}
\end{figure*}

\textbf{Unexpectedness.} Likewise, when comparing the results of the \textit{baseline} model (i.e., only the intercept) and the \textit{random intercept} model (which accounts for variations among participants), the results showed statistically significant variations in intercepts across participants. As a result, the random effects of the participants are included in the primary analysis. As shown in Fig. \ref{fig:q2-berkeley}, in general, participants perceive the recommendations as highly unexpected; participants in \textit{Exp} conditions (\textit{M} = 3.39, \textit{N} = 140) show similar level of unexpectedness about the recommendations compared to those in \textit{No-Exp} conditions (\textit{M} = 3.4, \textit{N} = 125). From the statistical analysis, the results show that the relationship between explanation and the unexpectedness of the course showed significant variance in intercepts across participants, \textit{SD} = 0.47 (95\% CI: 0.31, 0.72), $=\chi^2(1) = 9.905, p = .001$. However, the explanation has no significant effect on how surprisingly the participants perceive the course, \textit{b} = -0.014, \textit{t}(51)=0.074, \textit{p} = 0.94.

\begin{figure*}
  \begin{subfigure}{0.5\textwidth} 
    \centering
    \includegraphics[width=\linewidth]{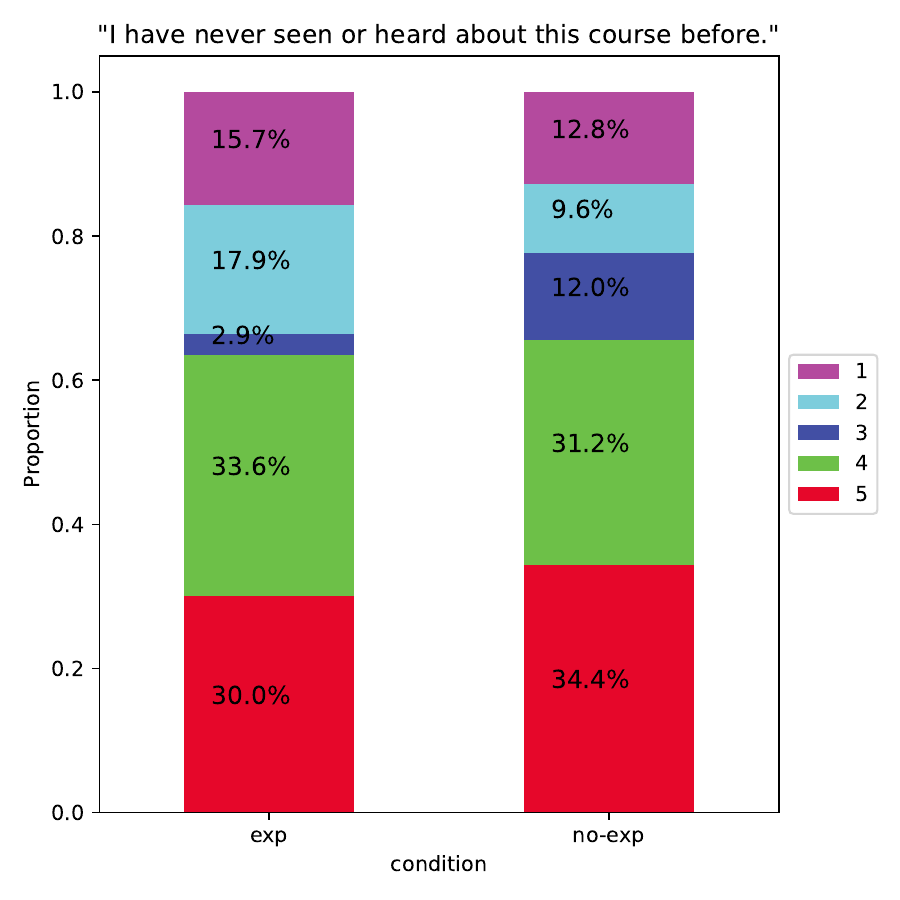}
    \caption{}
    \label{fig:q3-stacked-bar-berkeley}
  \end{subfigure}%
  \begin{subfigure}{0.5\textwidth} 
    \centering
    \includegraphics[width=\linewidth]{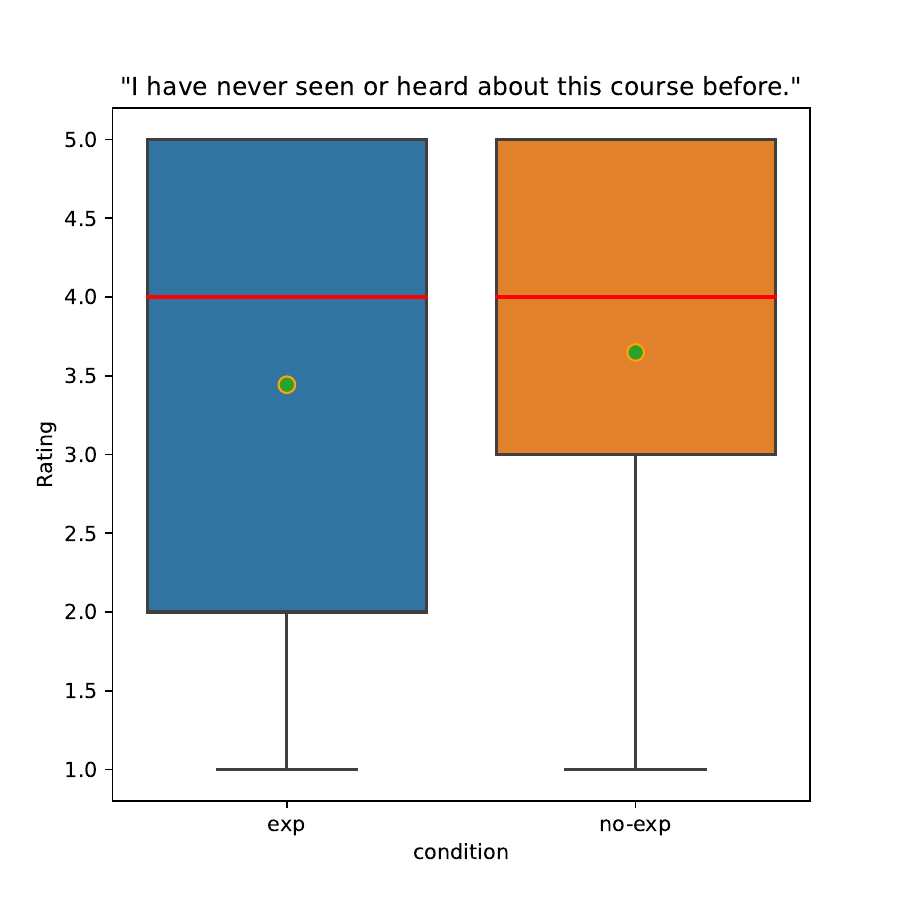}
    \caption{}
    \label{fig:q3-boxplot-berkeley}
  \end{subfigure}
  \caption{(a) Proportional distribution of user responses to the statement `I have never seen or heard about this course before.', comparing those with \textit{Explanation} (exp) and  \textit{Without Explanation} (no-exp). Ratings: 1 - `Strong Disagree', 2 - `Disagree', 3 - `Neutral', 4 - `Agree', 5 - `Strong Agree'. (b) A graph displaying the distribution of ratings in response to question Q3 for the two conditions, with the median indicated by red lines and the average represented by green circles.}
  \label{fig:q3-berkeley}
\end{figure*}

\textbf{Novelty.} Similarly, there are statistically significant variations in intercepts across participants, which were included in the primary analysis as a random effect. Fig. \ref{fig:q3-berkeley} shows that participants generally perceive the recommendations as highly novel, with an average rating of 3.54 out of 5. Participants in the \textit{Exp} conditions (\textit{M} = 3.44, \textit{N} = 140) perceive the recommended courses as slightly less novel compared to those in the \textit{No-Exp} conditions (\textit{M} = 3.65, \textit{N} = 125). From the statistical analysis, the results show that the relationship between explanation and the novelty of the course exhibits significant variance in intercepts across participants, \textit{SD} = 0.58 (95\% CI: 0.39, 0.86), $=\chi^2(1) = 12.46, p = .004$. Having explanations slightly decreases the level of novelty perceived by participants, but it is not statistically significant, \textit{b} = -0.205, \textit{t}(51)=0.902, \textit{p} = 0.371.

\begin{figure*}
  \begin{subfigure}{0.5\textwidth} 
    \centering
    \includegraphics[width=\linewidth]{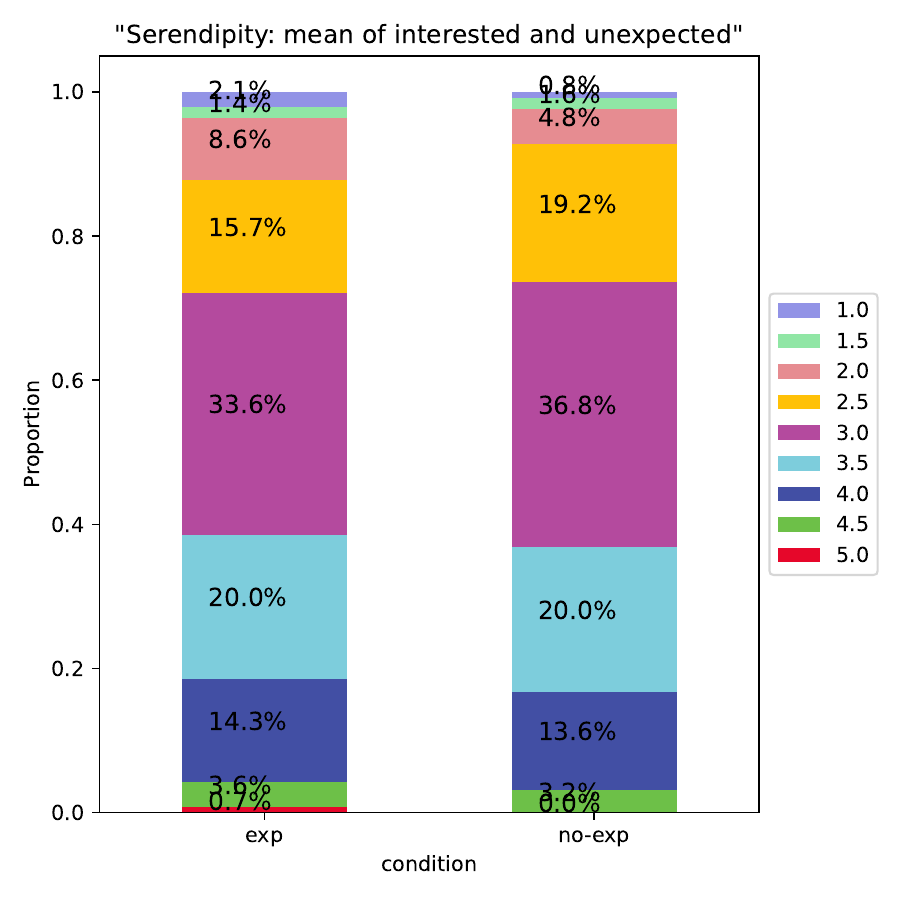}
    \caption{}
    \label{fig:q6-stacked-bar-berkeley}
  \end{subfigure}%
  \begin{subfigure}{0.5\textwidth} 
    \centering
    \includegraphics[width=\linewidth]{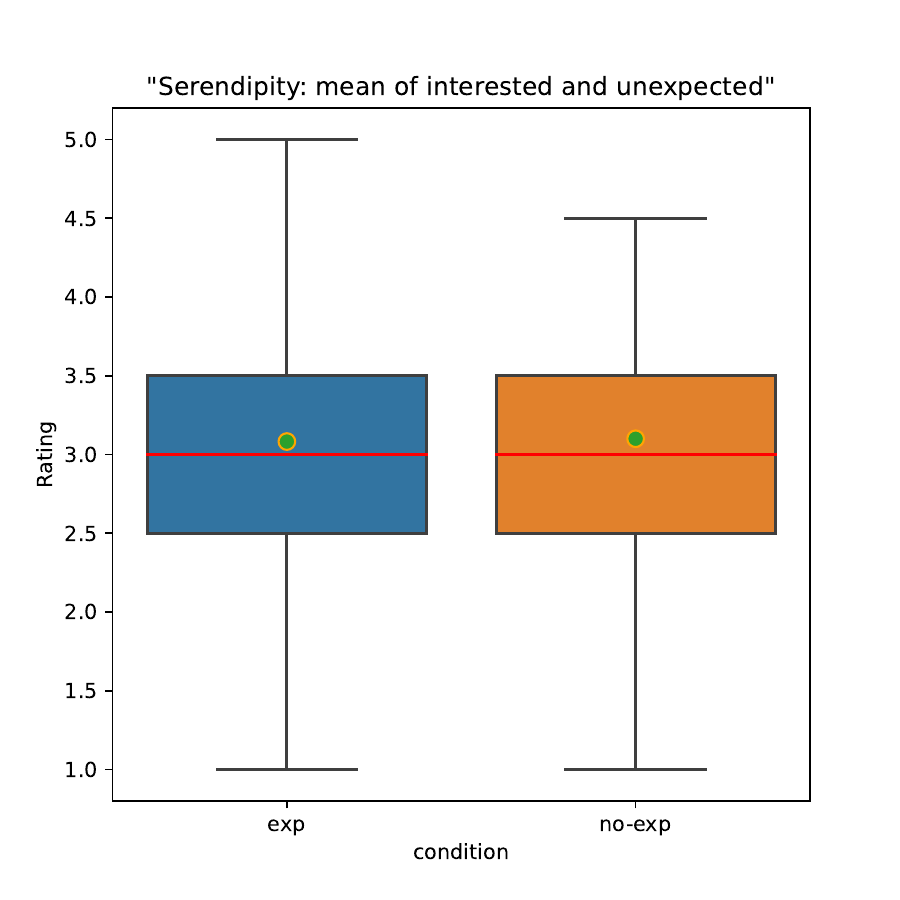}
    \caption{}
    \label{fig:q6-boxplot-berkeley}
  \end{subfigure}
  \caption{(a) Proportional distribution of the average ratings of questions Q1 and Q2 as a measure for serendipity. Original ratings of Q1 ad Q2: 1 - `Strong Disagree', 2 - `Disagree', 3 - `Neutral', 4 - `Agree', 5 - `Strong Agree'. (b) A graph displaying the distribution of the average ratings of questions Q1 and Q2 for the two conditions, with the median indicated by red lines and the average represented by green circles.}
  \label{fig:q6-berkeley}
\end{figure*}

\textbf{Serendipity.} Similar to the preliminary investigation described in \cite{Run@2021}, we evaluated serendipity by computing the mean of user-perceived unexpectedness and success \cite{shani2011evaluating,Run@2021}. In our primary analysis, we also factored in statistically significant variations in intercepts among participants, treating them as random effects. As depicted in Figure \ref{fig:q6-berkeley}, it's evident that, on the whole, participants in the \textit{Exp} conditions (\textit{M} = 3.08, \textit{N} = 140) exhibited a similar level of serendipity regarding the recommendations compared to those in the \textit{No-Exp} conditions (\textit{M} = 3.1, \textit{N} = 125). Our statistical analysis showed significant variance in intercepts among participants in the relationship between explanations and the level of serendipity of the course, \textit{SD} = 0.33 (95\% CI: 0.24, 0.46), $=\chi^2(1) = 21.20, p < .0001$. Yet, explanations had no impact on participants' perception of the serendipity of the course, with a coefficient of \textit{b} = -0.017, \textit{t}(51)=0.15, \textit{p} = 0.881.

Assessing serendipity poses a formidable challenge. While existing literature suggests that it can be approximated as the mean of user-perceived unexpectedness and success, it is imperative to acknowledge that highly unexpected items often yield lower perceived relevance. Our study supports this observation, as we found a notable negative relationship of -0.39 between relevance and unexpectedness, suggesting that subjects tend to prefer courses that are less likely to provide surprises. However, it is noteworthy that items characterized by both high unexpectedness and high relevance are the ones most valued by users.

\begin{figure*}
\centering
 \includegraphics[width=0.8\textwidth]{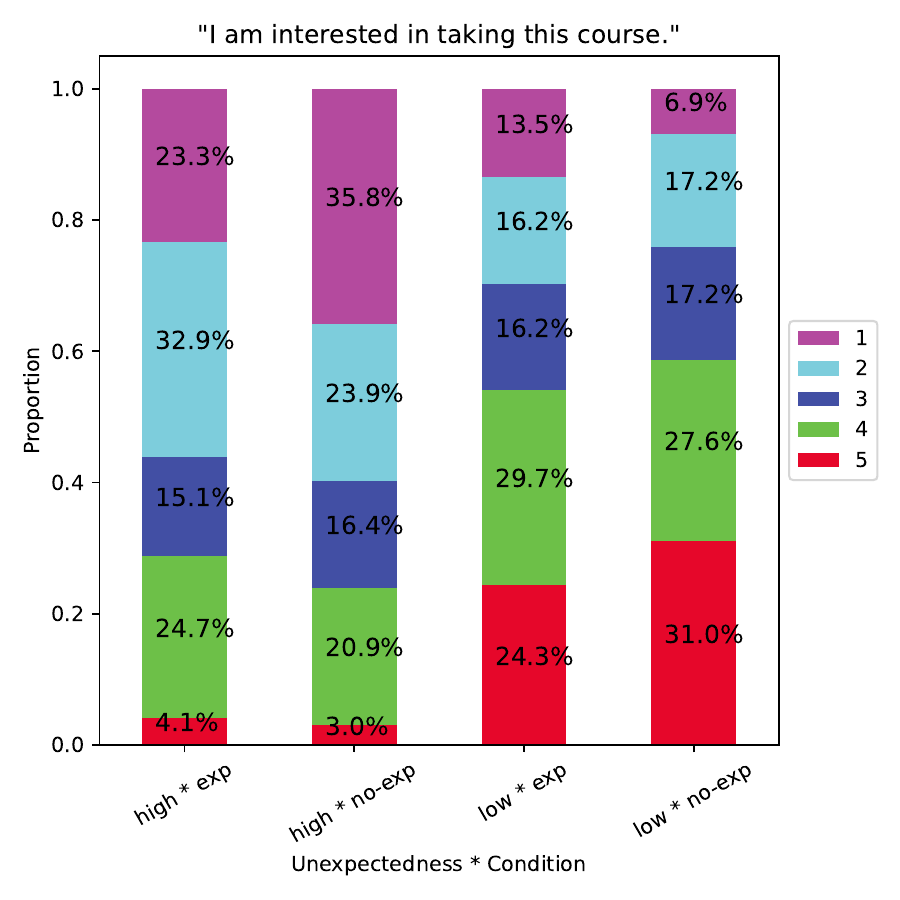}
 \caption{Proportional distribution of user responses to the statement `I am interested in taking this course.' across different \textit{Unexpectedness} levels and \textit{Explanation} conditions: \textit{High Unexpectedness} with \textit{Explanation} (high * exp), \textit{High Unexpectedness} without \textit{Explanation} (high * no-exp), \textit{Low Unexpectedness} with \textit{Explanation} (low * exp), and \textit{Low Unexpectedness} without \textit{Explanation} (low * no-exp). Ratings: 1 - `Strong Disagree', 2 - `Disagree', 3 - `Neutral', 4 - `Agree', 5 - `Strong Agree'.}
 \label{fig:q1_stacked_bar_unexpectedness}
\end{figure*}

To investigate further into the impact of explanations on user perception of recommendation interest across varying levels of unexpectedness, we categorized the 5-point Likert scale ratings of unexpectedness into `low' (comprising `Strongly Disagree' and `Disagree') and `high' (comprising `Strongly Agree' and `Agree') unexpectedness, with neutral ratings excluded from the analysis. As depicted in Figure \ref{fig:q1_stacked_bar_unexpectedness}, our findings indicate that, in comparison to courses with low unexpectedness, courses with high unexpectedness elicited reduced interest from participants. Specifically, when a course presented lower levels of unexpectedness, participants in the absence of explanations demonstrated a 0.234-point higher interest in pursuing the course. While this suggests that providing additional information about a course (including its potential knowledge benefits) may prompt subjects to realize its limited utility, this effect did not reach statistical significance (p-value = 0.445). Conversely, when a course exhibited high levels of unexpectedness, participants provided with explanations displayed a 0.22-point increase in their interest in taking the course. This effect also did not achieve statistical significance (p-value = 0.304). Nevertheless, it is noteworthy that explanations facilitated a more informed assessment of the course's utility and reduced the likelihood of its dismissal due to unfamiliarity.

\begin{figure*}
\centering
 \includegraphics[width=0.9\textwidth]{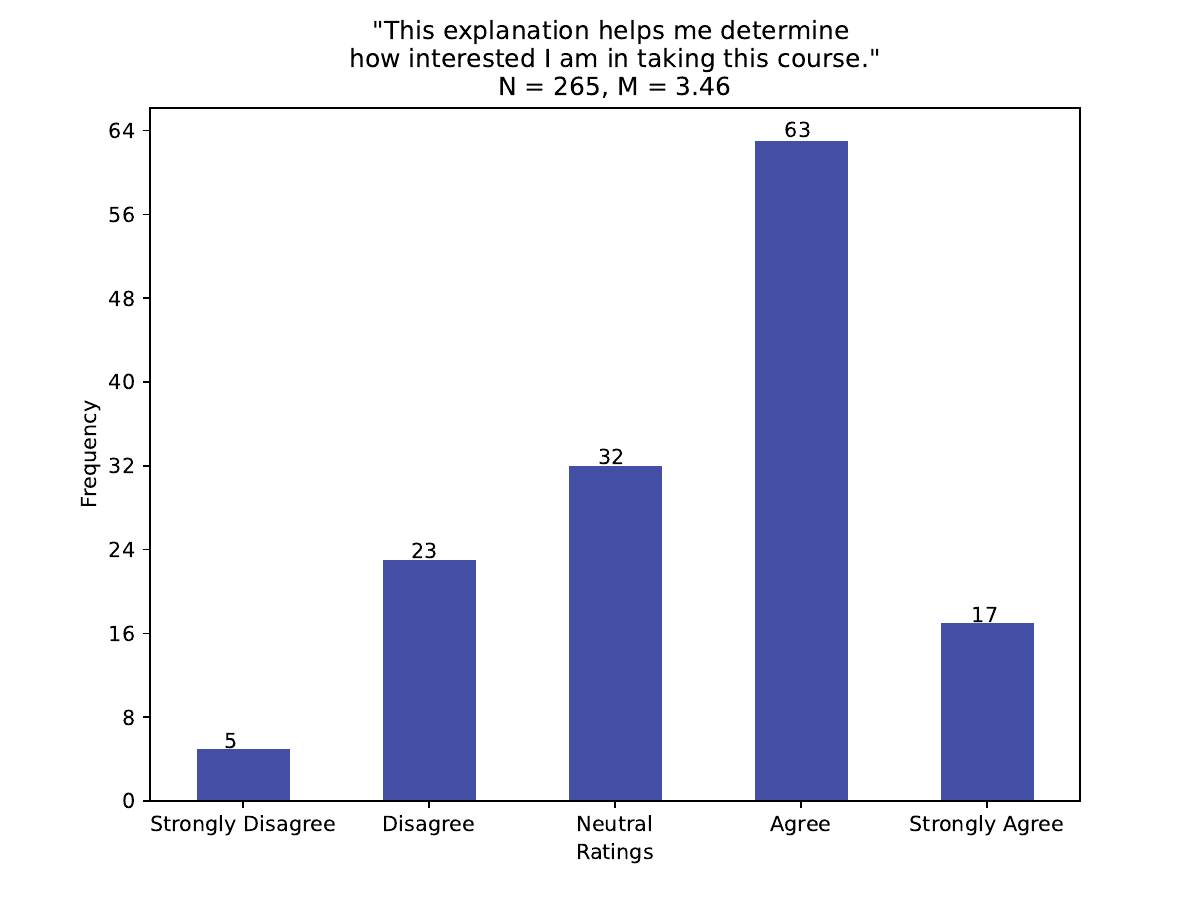}
 \caption{Frequency distribution of user responses to the statement `This explanation helps me determine how interested I am in taking this course.'.}
 \label{fig:q4_all_barplot}
\end{figure*}

\textbf{Explanation.} In accordance with the findings presented in Figure \ref{fig:q4_all_barplot}, it is evident that participants who were provided with explanatory information expressed a favorable disposition towards the utility of these explanations in influencing their interest in the recommendations (mean = 3.46, N = 265). Notably, a significant majority of respondents endorsed either `Agree' or `Strongly Agree' in response to this assertion. This implies that the provision of explanations serves as a valuable resource for participants, affording them a deeper understanding of the recommendations, which in turn facilitates informed decision-making. When being asked about how well the explanations assist participants in understanding the course's relevance to their field of study, it's clear that participants, as a whole, hold a fairly neutral position on this issue. The average rating of 3.04 out of 5 (N = 265) indicates a lack of strong agreement on whether the explanations effectively clarify how the course aligns with their academic interests.

\begin{figure*}
\centering
 \includegraphics[width=0.9\textwidth]{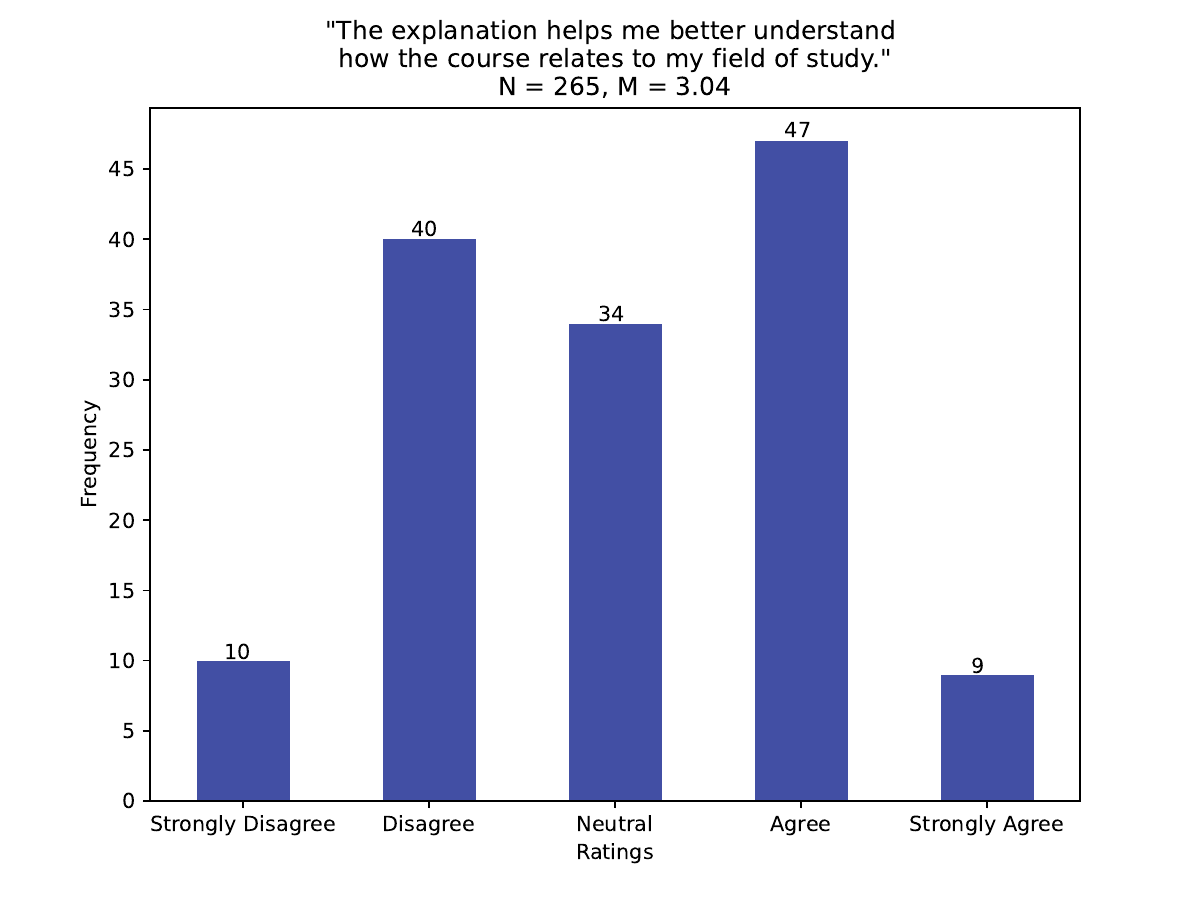}
 \caption{Frequency distribution of user responses to the statement `The explanation helps me better understand how the course relates to my field of study.'.}
 \label{fig:q5_all_barplot}
\end{figure*}

\subsubsection{A deeper analysis - Does explanation improve confidence in making decisions?} The hypothesis posits that when subjects are provided with explanations, they are better equipped to assess the utility of recommendations, feel more confident in their decision-making, and are less likely to give a neutral rating to an item. This would suggest that explanations play a pivotal role in guiding their choices by helping them discern their preferences. Additionally, as students who have not yet declared their majors (a majority in their early program) may possess less knowledge about courses and have more options to choose in comparison to those who have already declared their majors, we anticipate the presence of an interaction effect between the declaration of a major and the provision of explanations. Consequently, it is expected that explanations will have varying impacts on students with undeclared majors as opposed to those with declared majors. To achieve this, I first convert the original ratings (on a 5-point Likert scale) to neutral ratings (i.e., `Yes' for a rating of 3 (`Neutral') and `No' for all other ratings). 

\begin{figure*}
\centering
 \includegraphics[width=.9\textwidth]{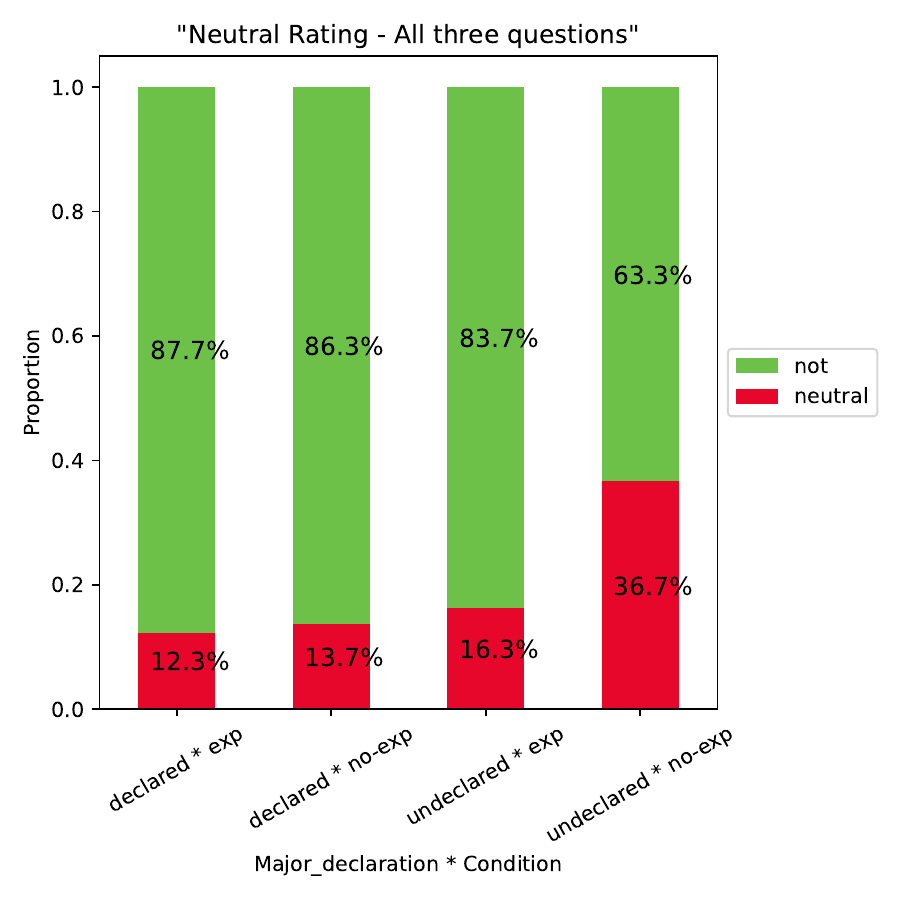}
 \caption{Distribution of `Neutral' ratings among four groups based on the interactions between major (declared vs. undeclared) and the presence of an explanation (vs. no explanation): declared * exp (N=285), declared * no-exp (N=285), undeclared * exp (N=135), undeclared * no-exp (N=90). The `Neutral' ratings are aggregated from the responses to the three primary research questions: Q1, Q2, and Q3. The percentage of `Neutral' ratings is 16.22\% (129 `Neutral' ratings of 795).}
 \label{fig:neutral_rating_declared}
\end{figure*}

There are 53 participants in our study, coming from diverse academic backgrounds (refer to Appendix \ref{appendix-course-rec}, Table \ref{tab:major_counts} for details). Out of this group, 15 individuals have not yet declared their majors. The distribution of `Neutral' ratings is displayed in Figure \ref{fig:neutral_rating_declared}, segmented into four distinct categories based on major declaration and the presence of an explanation. Our analysis reveals a compelling additive interaction effect between the declaration of a major and the provision of explanations. Specifically, subjects in the `Exp' groups consistently exhibit a reduced tendency to provide `Neutral' opinions. Similarly, individuals in the \textit{declared\_major} groups also display a decreased inclination toward `Neutral' ratings. Notably, those participants who have not yet declared their majors and do not receive explanations exhibit the highest percentage of `Neutral' ratings (36.7\%), significantly higher than those who receive explanations (16.3\%).

Our statistical analysis revealed a noteworthy interaction effect between the declaration of a major and the provision of explanations, yielding a p-value of 0.017. Upon further examination of the impact of explanations on participants belonging to declared and undeclared major groups, our findings showed that among participants with a declared major, the presence of explanations did not influence their neutral opinion significantly, as indicated by a p-value of 0.618. In contrast, for participants without a declared major, the absence of explanations was associated with a 0.20-point increase in their neutral opinion, and this difference was statistically significant, with a p-value of 0.0006.

We also examined the `Neutral' ratings for questions Q1 and Q2, as Q3 was excluded due to its minimal percentage of neutral responses. Notably, participants who have yet to declare their majors and who did not receive explanations showed the highest percentages of `Neutral' ratings: 40.0\% for Q1 and 42.2\% for Q2 (refer to Appendix \ref{appendix-course-rec}, Fig. \ref{fig:a1_neutral_rating} and \ref{fig:a2_neutral_rating}). For Q1, the data revealed that participants without a declared major were 0.20 points more likely to choose a neutral stance in the absence of explanations. This trend approached statistical significance with a p-value of 0.06, possibly affected by a smaller sample size. For Q2, these participants exhibited a 0.14-point increase in neutrality without explanations, although it was not significant with a p-value of 0.2. Overall, for both questions, the absence of explanations led those without a declared major to be more inclined to rate as `Neutral.' This was statistically significant with p-values of 0.01 for Q1 and 0.003 for Q2, respectively.

\section{Discussion, Limitations \& Future Work}

This study has limitations arising from methodology, resources, and current technology. Recognizing these helps clarify the scope and guide future work. Below, we summarize key limitations and directions.

\textbf{Automatic knowledge extraction.}
Skills are central to educational AI, aiding both recommendations and explanations. However, the absence of a standardized higher-education/job-market knowledge base limits progress. We developed and evaluated a concept extraction model for educational documents, demonstrating strong performance, but this is only an initial step toward automated educational ontology construction. Ontology building also requires relation extraction, hierarchy creation, and concept disambiguation \citep{shamsfard2004learning, Wilson@2012}. Large Language Models (LLMs) \cite{NEURIPS2020_1457c0d6,NEURIPS2022_b1efde53}, especially when instruction-tuned \cite{zhou2023universalner,wang2023instructuie,agrawal-etal-2022-large}, offer promising capabilities for these tasks. Further work should explore LLM-based pipelines for full ontology automation.

\textbf{Skill-based explanation for course recommendation.} Our study focuses on enhancing course recommendation systems in higher education by incorporating skill-based explanations. Findings from our first study reveal several limitations that warrant attention in future research.

Although PLAN-BERT effectively leverages past enrollment sequences and user/item features to generate recommendations, our current diversification strategy—limiting results to one course per department—is overly simplistic. This rule may introduce irrelevant suggestions, as some departments are more closely related to other disciplines, while others have fewer natural connections. A more refined approach would replace the rigid “one-per-department” constraint with a relevance threshold. Departments failing to meet this threshold could then be bypassed, allowing multiple recommendations from the same department when appropriate.

Another promising direction is to frame recommendation diversity and serendipity as a multi-task or multi-label optimization problem during training \cite{Svore2011,Mahapatra2023}. This would enable the system to jointly optimize for both relevance and unexpectedness, producing balanced, non-dominated solutions. Implementing such an approach would require collecting labeled data on the “unexpectedness” of relevant courses to train the models effectively.

While this study did not conclusively determine whether explanations improve recommendation effectiveness, it suggests that offering detailed rationales—highlighting why a course is recommended and how it aligns with a student’s abilities and interests—could increase the perceived value of recommendations. The impact of explanations may vary by academic discipline and stage of study. Notably, we observed a positive effect on students’ interest in unexpected courses, especially among those without a declared major. Future work should focus on this demographic with larger sample sizes to validate these findings.

Our current system also does not fully account for course level, sequence, or prerequisite structure, sometimes producing suboptimal recommendations. Integrating prerequisite and course-sequence information could substantially improve quality. Advanced models, including deep sequential or Transformer-based architectures, could infer course ordering patterns from historical enrollment data. Additionally, considering advanced placement credits and equivalent courses would help avoid recommending content the student has effectively already completed.

A further limitation is the sample size in our between-subjects experiments—46 participants in the first study and 53 in the second—which may have reduced statistical power. Larger-scale studies would yield more robust conclusions. Moreover, our work has focused primarily on explanations for improving recommendation accuracy. Future research should explore other dimensions of explanation utility, such as transparency, trust, persuasiveness, and user satisfaction \cite{Tintarev2007,tintarev2015explaining,Zhang2020}, to develop a broader understanding of their role in recommender systems.

Lastly, our reliance on questionnaire-based user feedback comes with inherent drawbacks. Preferences expressed in surveys may be shaped by mood, bias, social desirability, or limited self-awareness, leading to incomplete or misleading data. In some cases, students may report liking a course but ultimately choose not to enroll. A more reliable evaluation could track actual enrollment behavior, though this approach raises privacy concerns and is subject to other confounding factors such as scheduling conflicts or course demand. Despite these challenges, exposing students to relevant and engaging course options remains valuable, as it supports informed decision-making and broadens academic exploration.

\textbf{Large language models.}
This paper applies NLP methods across document representation, knowledge extraction, recommendation, and explanation generation. Large Language Models (LLMs), such as ChatGPT, have transformed these domains through their broad generalization abilities, achieved via multi-task training and unified encoding. Leveraging effective prompting, instruction fine-tuning, reinforcement learning from human feedback (RLHF) and LLM-generated embeddings \cite{lin2023vector,mialon2023augmented} can enhance skill detection, relationship extraction, and overall system explainability in educational information systems.

LLMs can be integrated into intermediate stages to improve downstream tasks or directly applied to produce recommendations and explanations. They also hold promise for educational ontology development, from dataset curation to relation extraction. Such capabilities could refine skill diversification and yield more nuanced, skill-focused recommendations.

As LLMs evolve, they are poised to play an increasingly central role in advancing NLP applications in education and beyond.

\backmatter








\begin{appendices}

\section{Course Recommendation}
\label{appendix-course-rec}

\begin{table}[h!]
    \centering
    \small
    \caption{Summary of Majors of Participants.}
    \begin{tabularx}{\textwidth}{|X|c|}
    \hline
    \textbf{Majors} & \textbf{Number of Subjects} \\
    \hline
    `Letters \& Sci Undeclared' & 15 \\ \hline
    `Molecular \& Cell Biology' & 3 \\     \hline
    `Mechanical Engineering' & 3 \\    \hline
    `L\&S Computer Science' & 3 \\    \hline
    `Electrical Eng \& Comp Sci' & 2 \\    \hline
    `Chemistry' & 2 \\    \hline
    `L\&S Public Health' & 2 \\    \hline
    `Bioengineering' & 2 \\    \hline
    `Industrial Eng \& Ops Rsch' & 1 \\    \hline
    `Molecular Environ Biology' & 1 \\    \hline
    `Media Studies' & 1 \\    \hline
    `Mathematics' & 1 \\    \hline
    `L\&S Data Science' & 1 \\    \hline
    `Integrative Biology' & 1 \\    \hline
    `Info \& Data Science-MIDS' & 1 \\    \hline
    `Applied Mathematics', `L\&S Computer Science' & 1 \\    \hline
    `Engineering Physics' & 1 \\    \hline
    `Economics' & 1 \\    \hline
    `Economics', `L\&S Data Science' & 1 \\    \hline
    `Economics', `L\&S Data Science', `L\&S Ops Research \& Mgmt Sci' & 1 \\    \hline
    `Economics', `French', `History', `Philosophy' & 1 \\    \hline
    `Economics', `Electrical Eng \& Comp Sci', `L\&S Data Science' & 1 \\    \hline
    `EECS/MSE Joint Major' & 1 \\    \hline
    `Cognitive Science' & 1 \\    \hline
    `Cognitive Science', `L\&S Computer Science' & 1 \\    \hline
    `Civil Engineering' & 1 \\    \hline
    `Business Administration' & 1 \\    \hline
    `Business Administration', `Electrical Eng \& Comp Sci' & 1 \\    \hline
    `Statistics' & 1 \\
    \hline
    \end{tabularx}
    \label{tab:major_counts}
\end{table}

\begin{figure}
\centering
 \includegraphics[width=.8\textwidth]{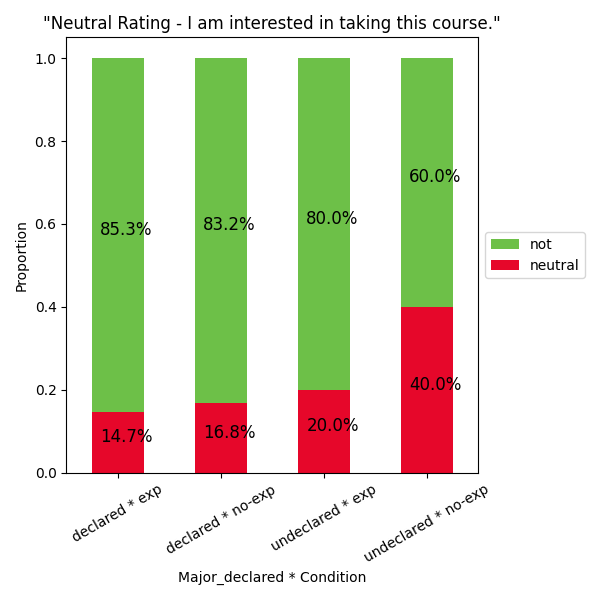}
 \caption{Distribution of `Neutral' ratings for the statement `I am interested in taking this course.' among four groups based on the interactions between major (declared vs. undeclared) and the presence of an explanation (vs. no explanation): declared * exp (N=95), declared * no-exp (N=95), undeclared * exp (N=45), undeclared * no-exp (N=30). The `Neutral' ratings are aggregated from the responses to the three primary research questions: Q1, Q2, and Q3. The percentage of `Neutral' ratings is 19.24\% (51 `Neutral' ratings of 265).}
 \label{fig:a1_neutral_rating}
\end{figure}

\begin{figure}
\centering
 \includegraphics[width=.8\textwidth]{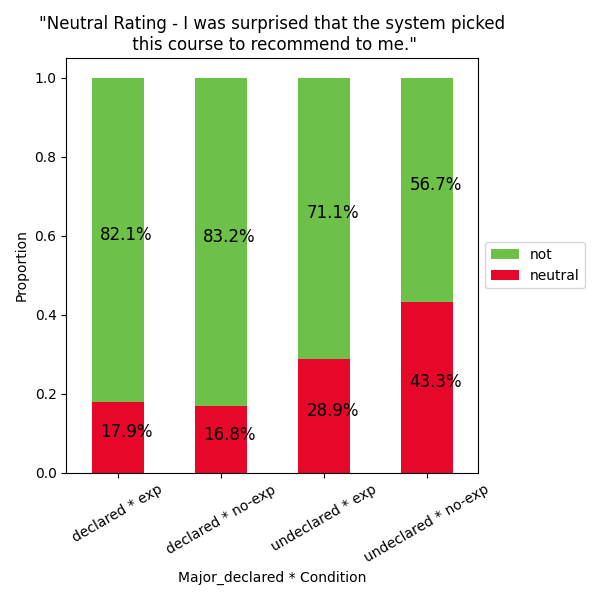}
 \caption{Distribution of `Neutral' ratings for the statement `I was surprised that the system picked this course to recommend to me.' among four groups based on the interactions between major (declared vs. undeclared) and the presence of an explanation (vs. no explanation): declared * exp (N=95), declared * no-exp (N=95), undeclared * exp (N=45), undeclared * no-exp (N=30). The percentage of `Neutral' ratings is 22.26\% (59 `Neutral' ratings of 265).}
 \label{fig:a2_neutral_rating}
\end{figure}




\end{appendices}

\bibliographystyle{plain}
\bibliography{sn-article}

\end{document}